\documentclass[conference]{IEEEtran}

\usepackage{amsmath,amscd,amssymb,amsgen,amsfonts,amsbsy}
\usepackage{mathrsfs}
\usepackage[utf8x]{inputenc}
\usepackage{color}
\usepackage{textcomp}
\usepackage{float}
\usepackage{latexsym,graphicx}
\usepackage{mathtools}
\usepackage{breqn}
\usepackage{tikz}
\usepackage{breqn}
\usetikzlibrary{automata,arrows,positioning,calc}
\usepackage{ragged2e}
\usepackage{url}
\usepackage{cite}
\usepackage{algorithmic}
\usepackage[ruled,lined]{algorithm2e}
 \usepackage{blkarray}

\newcommand{\remove}[1]{}

\newcommand{\comments}[1]{}

\newcommand{\expect}[1]{\mathsf{E}\left({#1}\right)}

\newtheorem{lemma}{Lemma}
\begin{document}


\title{Whittle Index Learning  Algorithms for  Restless Bandits with Constant Stepsizes 
}

\author{
 Vishesh Mittal \ \ \ & \ \ \ Rahul Meshram \ \ \ & \ \ \ Surya Prakash  \\
ECE Deptt. \ \ \ &   \ \ \ EE Deptt.  \ \ \  & \ \ \ ECE Deptt.  \\ 
IIIT Allahabad \ \ \ & \ \ \ IIT Madras  \ \ \ & \ \ \ IIIT Allahabad}
\maketitle

\begin{abstract}
    We study the Whittle index learning algorithm for restless multi-armed bandits. We consider index learning algorithm with Q-learning. We first present Q-learning algorithm with exploration policies---$\epsilon$-greedy, softmax, $\epsilon$-softmax with constant stepsizes. We extend the study of Q-learning to index learning for single-armed restless bandit. 
    The algorithm of index  learning is two-timescale variant of stochastic approximation, on slower timescale we update index learning scheme and on faster timescale we update Q-learning assuming fixed index value. In Q-learning updates are in asynchronous manner. 
    We study constant stepsizes two timescale stochastic approximation algorithm. 
    We provide analysis of two-timescale stochastic approximation for index learning with constant stepsizes.  
    Further, We present study on index learning with deep Q-network (DQN) learning and linear function approximation with state-aggregation method. 

    We describe the performance of our algorithms using numerical examples. We have shown that index learning with Q learning, DQN and function approximations learns the Whittle index.

\end{abstract}

\section{Introduction}

Restless multi-armed bandit (RMAB)  has applications to resource allocation problems---opportunistic communication systems \cite{Borkar17b}, \cite{Xiong2023}, queueing systems \cite{Verloop16,Glazebrook05}, machine repair and maintenance \cite{Glazebrook2006}, healthcare \cite{Killian2021,Mate21}, recommendation systems \cite{Meshram15,Meshram2016,Meshram17}, and multi-target tracking systems  \cite{Ny2008,Whittle88}. Other applications of RMAB are given in \cite{Gittins11,NinoMora23}. 

Recently, in the context of semantic communications, RMAB can be describe the model of resource allocation in a network where multiple communication channels can be scheduled dynamically/ The goal is to maximize the relevant semantic information transmitted over channels, rather than the raw data throughput. In semantic communication, the relevance and the impact of transmitted message are more important than simply transmitting the data with high fidelity. We now describe restless bandits in semantic communications. Arms of RMAB represent communication channels or different users/devices in networks. We assume that there arms are independent. Each arm can be in different state, that represent quality of channels (low or high). Whenever arm is pulled, there is a reward, it is  measure of the successful transmission of relevant information. State of each arm evolves according to Markov process, regardless of whether the arm is pulled. This evolution can model the changing network conditions, user mobility etc. The objective is to develop a policy that selects which arms (communication channels) to use at each time step to maximize the long term cummulative semantic rewards.

RMAB consists of $N$ independent arms, i.e, independent Markov processes and state of arms is evolving at each time step. The agent has to determine at each time step which sequence to act or play an arm so total expected cumulative reward is maximized. 

RMAB problem first introdued in \cite{Whittle88}, and  studied Whittle index based policy and it is shown to be near optimal,  \cite{Weber90}.  In the Whittle index policy, an arm with the highest index based on the state of each arm is played at each time step. To use the Whittle index based policy, the agent need to know the model, i.e., transition probabilities from a state to another state and reward matrix. In this paper we assume that this information is not available. The agent employ learning algorithms. 

There are many  learning algorithms for RMAB are studied see \cite{Tekin2012,Meshram17,Fu2019,Jung2019,Wang2020,Nakhleh2021,Akbarzadeh2022}.  Thompson sampling (TS), upper confidence bound (UCB), and variant of TS and UCB algorithms are studied for RMAB, where  objective is to consider parameterized or estimate the model and it is formulated as regret minimization problem. The regret computation depends on structural of the model and it require stronger assumptions on prior or mixing time to obtain regret bounds.  Other learning approach include Q-learning algorithm and it is model-free, where the objective is  to learn the index of each arm for given state, \cite{Fu2019,Avrachenkov2022}. 



Q-learning is a well studied algorithm for Markov decision processes (MDPs) when the model is unknown to the agent, \cite{Watkins1989,Borkar2000,Borkar08,Sutton2018}. The convergence analysis of Q-learning algorithm is provided using stochastic approximation scheme, \cite{Satinder2000,Borkar2000}. Here, the convergence depends on the frequency of update of each state-action pair, Further, the rate of convergence depends on structure of model and an exploration policy that is employed for action selection during Q-learning updates, for examples---$\epsilon-$greedy, and softmax policy.    


In this paper, we discuss Q learning and different action selection policies---$\epsilon$-greedy, softmax and $\epsilon$-softmax. We next study the index learning algorithm using Q-learning  with constant stepsizes and action selection policies. We further consider deep Q learning network (DQN) and linear  function approximation approach. We illustrate the significance of action-selection policy for faster convergence of algorithm.

\subsection{Related work} 

Q-learning algorithm for discounted MDP is first introduced in \cite{Watkins1989}. The convergence analysis of Q-learning algorithm  is  discussed \cite{Jaakkola1993,Satinder2000,Borkar08,Sutton2018}. 
In \cite{Sutton2018}, various reinforcement learning algorithms are discussed for Markov decision processes (MDPs).   Deep-Q-Network (DQN)  based learning algorithm is studied in \cite{Mnih2015} which uses non-linear function approximation with deep neural network architecture.

Literature on Q-learning algorithm for restless bandits is limited, \cite{Fu2019,Robledo2021,Killian2021,Avrachenkov2022, Xiong2023}. Neural-network based model is proposed to learn Whittle index for restless bandits in \cite{Nakhleh2021}.  The neural networks approach is combined with Q-learning in \cite{Robledo2022} and they studied QWI and QWINN. 
The analysis of Q-learning algorithm is based on two-timescale stochastic approximations with decreasing stepsizes.














\subsection{Our contributions}
Our contributions are as follows. 
We consider Q-learning algorithm for a single-armed restless bandit (SARB). We discuss the Q-learning algorithm for MDP with action selection policies. We extend Q-learning algorithm for index learning. It is two-timescale algorithm with constant stepsizes--- faster timescale  Q-learning is performed and slower scale index is updated. Stepsizes are constant in both algorithms.  The analysis is discussed with two timescale constant stepsizes approximations. 
The exploration scheme $\epsilon$-greedy scheme do not give always faster convergence. It motivate us to look for alternative exploration schemes---softmax and $\epsilon$ softmax.  This is not considered in preceding works.

The convergence of index algorithm depends on Q-learning scheme, this convergence depends on model and it can be slow. To accelerate convergence of learning learning algorithm, we use  we use re-initialization approach after periodic interval which can ensures that all state action pair sufficiently visited. We also study different exploration strategies using action selection method.

Further, we study DQN based index learning algorithm and  function approximation based index learning. In the  function approximation based index learning, we propose state-aggregation approach. We present the numerical examples to illustrate the performance of proposed scheme.  

We also present comparison of  computational time for index learning with Q-learning and DQN method. 

Paper is organized as follows. In section \ref{sec:preliminaries}, we discuss preliminaries on Q-learning. We study index learning algorithm with Q-learning approach in \ref{sec:SARB-index-Q-learning}. We also present DQN algorithm for index learning in \ref{sec:index-learning-DQN}. We demonstrate performance of proposed learning algorithms using numerical example in section \ref{sec:numerical examples} and concluding remarks in section \ref{sec:conclusion}.

\section{Preliminaries on MDP and Q Learning}
\label{sec:preliminaries}
Consider an agent interact with an environment through sequence of actions. The goal of agent  is to maximize the long term cumulative discounted reward function

\begin{equation}
   V^{*}(s) = \max_{\pi} \mathbb{E}\left[ \sum_{t=0}^{\infty} \beta^t r_t \vert s_0 =s,  \pi  \right]
\end{equation}

We assume the finite state and finite action MDP. Let state $x_n \in \mathcal{S},$  action of the agent $a_n \in \mathcal{A}$ and agent received reward $r_n = r(x_n,a_n)$ at time step $n.$ The state of an environment changes to $x_{n+1} = s^{\prime}$ from state $x_n=s$ under action $a_n =a$ according to $p_{i,j}^a.$ The transition probability matrix is given by $P^a = [[p_{i,j}^a]].$ The optimal policy is stationary and an optimal dynamic program is given by 
\begin{eqnarray*}
    Q^*(s,a) &=& r(s,a) + \beta \sum_{s^{\prime} \in \mathcal{S} } p_{s,s^{\prime}}^a  
   \max_{a^{\prime}} Q^*(s^{\prime},a^{\prime}), 
 \\
V^{*}(s) &=& \max_{a \in \mathcal{A}} Q^*(s,a).
\end{eqnarray*}

We assume that reward matrix $R =[[r(s,a)]]$ and probability matrix $P^a,$ $a \in \mathcal{A}$ are unknown. In \cite{Watkins1989}, Q-learning algorithm is proposed which an incremental learning scheme. Here,  the agent experience the  data: in the $n^{th}$ time step, the agent  observes its current state $x_{n} = s,$ selects  an action $a_n = a,$ 
observes the next state $x_{n+1} = s^{\prime},$ and receives an immediate reward $r_n = r(s,a).$ Then agent updates its estimate on Q-function (state-action value) at time step $n+1$ using previous Q-value estimate $Q_{n}$ according to: 
\begin{eqnarray}
Q_{n+1}(s, a) = Q_{n}(s, a) + \alpha_n \left[r_n + \beta  \max_{a^{\prime}} Q_{n}(s^{\prime},a^{\prime}) 
\right. \nonumber \\ \left.
- Q_{n}(s, a)  \right] 
\end{eqnarray}
for $x_n=s,$ and $a_n = a,$ 
$(s,a) \in \mathcal{S} \times \mathcal{A},$ otherwise
$Q_{n+1}(s,a) = Q_n(s,a).$
Here,  $\alpha_n$ is learning rate. 
This is also referred as asynchronous Q-learning, only one state-action pair is updated at a given time and updates are asynchronous. 
From \cite{Watkins1989,Jaakkola1993,Borkar08}, Q-learning algorithm converges to desired Q-value function if  for each state-action $(s,a) \in \mathcal{S} \times \mathcal{A},$ the $Q_n(s,a)$ estimate are updated infinitely often as $n \rightarrow \infty.$  The convergence  result is given in the following lemma.
\begin{lemma}
Suppose 
\begin{itemize}
    \item  $\alpha_n \rightarrow 0$ as $n \rightarrow \infty$
    \item $\sum \alpha_n = \infty$ and $\sum \alpha_n^2 < \infty$
    \item $R_{\max} = \max_{a \in \mathcal{A}}\max_{s \in \mathcal{S}}r(s,a) < \infty $
    \item For each state-action pair $(s,a),$  $Q_n(s,a)$ is updated infinitely often as $n \rightarrow \infty.$
\end{itemize}
Then $Q_n(s,a) \rightarrow Q^*(s,a)$ for all $(s,a)$ almost surely.  
\end{lemma}

The proof is using stochastic approximation scheme with ODE approach. 
The limiting o.d.e, is given by 
\begin{eqnarray*}
    \dot{Q} = F(Q)- Q,
\end{eqnarray*}
where $F(Q) = r(s,a) + \beta \sum_{s^{\prime} \in \mathcal{S} } p_{s,s^{\prime}}^a  
   \max_{a^{\prime}} Q(s^{\prime},a^{\prime}).$ 
The Q-learning update $Q_n$ asymptotically tracks solution of the limiting o.d.e. Here, $F$ is contraction map w.r.t. max norm $||\cdot||_{\infty}.$  It has global asymptotically stable unique equilibrium.
The  proof  is discussed in  \cite{Borkar2000}.  

The Q-value estimates are improved in each step using the one-step TD error $r + \gamma  \max_{a'} Q(s', a') - Q(s, a).$   
The one-step TD error is weighted by the step-size to facilitate stable learning. A larger step-size can lead to faster initial learning, but could face difficulties in convergence later. In contrast, a smaller step-size will lead to slower initial learning, but would lead to more optimal performance, since the algorithm will be able to adjust the estimates with a greater precision.
Q-learning is described in Algorithm~\ref{algo:I}.

\begin{algorithm}
    \KwIn{Step-size, $\alpha \in {(0, 1]}$; small $\epsilon > 0$; discount factor, $\gamma$; threshold for convergence, $\delta$; maximum number of iterations, $T_{max}$}
    \KwOut{Q value estimate $Q(s, a)$}

    \textbf{Initialize}: $Q(s, a)$ for all $s \in \mathcal{S}, a \in \mathcal{A}(s)$ 
    \vspace{10pt}

    \For {$n=0, 1, 2, \cdots, T_{\max}-1 $ }
    {
        For current state $s$ choose $a$  according to exploration policy (e.g., $\epsilon$-greedy, softmax, $\epsilon$-softmax) \\
        Take action $a$, observe $r, s'$ \\
        $\mathrm{err} = r + \gamma \hspace{2pt} \max_{a'} Q(s', a') - Q(s, a)$ \\
        $Q(s, a) \leftarrow Q(s, a) + \alpha \times  \mathrm{err}$ \\
        $s \leftarrow s'$

        \vspace{5pt}
        \If {$\alpha \times \mathrm{err} < \delta$}{
            $\mathrm{break}$
        }
    }
 
    \vspace{10pt}
    \Return{$Q(s, a)$} \;
    \caption{Q-learning algorithm}
    \label{algo:I}
\end{algorithm}

\subsection{Q-Learning with different exploration schemes}
We now discuss different exploration schemes. 
\subsubsection{$\epsilon-$greedy exploration scheme}

The action is selected corresponding to the current state using $\epsilon-$ greedy policy, at time step $n$ with state $s$. It is given by 
\begin{eqnarray*}
    a_n =  
    \begin{cases}
        a & \mbox{random action $a$ with prob. $\frac{\epsilon}{|\mathcal{A}|}$} \\
        \arg\max_{a \in \mathcal{A}} Q_{n+1}(s,a) & \mbox{with prob. $1-\epsilon.$}
    \end{cases}
\end{eqnarray*}

\subsubsection{Softmax exploration scheme}
Here, action is selected according to softmax (Boltzman) exploration rule as follows. 
\begin{eqnarray}
    Pr(a|s, n-1, Q_{n-1}) = \frac{e^{Q_{n-1}(s,a)}}{\sum_{b \in A}e^{Q_{n-1}(s,b)}},
\end{eqnarray}
%
\subsubsection{$\epsilon-$softmax exploration scheme}
Inspired from $\epsilon$-greedy scheme, we also study $\epsilon$-softmax exploration scheme for action selection. 
\begin{eqnarray}
    a_n =  
    \begin{cases}
        a & \mbox{random action $a$ with prob. $\frac{\epsilon}{|\mathcal{A}|}$} \\
        \frac{e^{Q_{n-1}(s,a)}}{\sum_{b \in A}e^{Q_{n-1}(s,b)}} & \mbox{with prob. $1-\epsilon.$}
    \end{cases}
\end{eqnarray}
%
%
\section{A single armed restless bandit model and Index Learning}
\label{sec:SARB-index-Q-learning}
We extend Q-learning study to index learning for a single-armed restless bandit (SARB). In this model, we assume that it is finite states $\mathcal{S}$ and two-actions MDP $\mathcal{A} = \{0,1\}.$ There is a subsidy $\lambda$ for action $a =0.$ The dynamic program is given by 
\begin{eqnarray*}
    Q^{\lambda}(s,a) =  r(s,a) + (1-a)\lambda +   \beta \sum_{s^{\prime} \in \mathcal{S} } p_{s,s^{\prime}}^a  
   \max_{a^{\prime}} Q^{\lambda}(s^{\prime},a^{\prime}),
\end{eqnarray*}
\begin{eqnarray*}
    V(s) &=& \max_{a \in \mathcal{A}} Q(s,a).
\end{eqnarray*}
The Whittle index is the minimum subsidy requires to hold the state-action value function to be equal for both actions, that is, there is $\lambda$ such that $Q^{\lambda}(s,a=1) - Q^{\lambda}(s,a=0) =0.$ 
The index $W(s) = \min \{ \lambda:  Q^{\lambda}(s,a=1) - Q^{\lambda}(s,a=0) =0 \}$ for $s \in \mathcal{S}.$ This is applicable when model is known.
Our interest here is to study Q-learning algorithm for fixed subsidy $\lambda$  and it is as follows.  
\begin{eqnarray}
Q_{n+1}^{\lambda}(s, a) = Q_{n}^{\lambda}(s, a) + \alpha_n \left[r(s,a) + \lambda (1-a) + 
\right. \nonumber \\ \left. 
\beta  \max_{a^{\prime}} Q_{n}^{\lambda}(s^{\prime},a^{\prime}) 
- Q_{n}^{\lambda}(s, a)  \right] 
\end{eqnarray}
for $x_n=s,$ and $a_n = a,$ 
$(s,a) \in \mathcal{S} \times \mathcal{A},$ otherwise
$Q_{n+1}^{\lambda}(s,a) = Q_n^{\lambda}(s,a).$  Here $\alpha_n$ is learning rate (stepsizes). Suppose the index (subsidy) is known to decision maker but the model is unknown, then running Q-learning algorithm one can learn the optimal Q-state value function. This is equivalent to solving Q-learning algorithm for MDP.
\subsection{Two-timescale index learning algorithm}
In this section, we assume that index $\lambda$ is not known the decision maker.
The index is function of state $\tilde{s},$  thus one requires to learn index $\lambda$ for each $\tilde{s} \in \mathcal{S}.$ We define $Q_n(s, a, \tilde{s})$ estimate at time $n$ for  $(s,a,\tilde{s}),$ where $\tilde{s}$ is a threshold state such that  $Q_n^{\lambda(\tilde{s})}(\tilde{s},a = 1, \tilde{s}) - Q_n^{\lambda(\tilde{s})}(\tilde{s},a = 0, \tilde{s}) = 0.$ 
\begin{algorithm}
    \KwIn{Step-sizes, $\alpha_1$ and $\alpha_2 \in {(0, 1]}$; small $\epsilon > 0$; discount factor, $\gamma$; threshold for convergence, $\delta$; maximum number of iterations, for step-1, $T_{\max}$ and for agent, $K_{\max}$}
    \KwOut{Indices, $\lambda(s)$}

    \textbf{Initialization}: $Q(s, a, \tilde{s})$ for all $s, \tilde{s} \in \mathcal{S}, a \in \mathcal{A}(s)$ arbitrarily; $\lambda(s)$ for all $s \in \mathcal{S}$; $s$ from  $\mathcal{S}$ arbitrarily
    \vspace{10pt}

    \For {$k = 0, 1, 2, \cdots, K_{\max}-1 $}
    {
        \For {$\tilde{s} = 0, 1, \cdots, S - 1$}
        {
            $q(\cdot , \cdot) \leftarrow Q(\cdot , \cdot , \tilde{s})$
            
            \For {$n = 0, 1, 2, \cdots, T_{\max}-1 $}
            {
                Choose $a$ from $s$ using policy derived from $Q$ (e.g., $\epsilon$-greedy) \\
                Take action $a$, observe $r, s'$ \\
                $\mathrm{err} = r + (1 - a) \lambda(\tilde{s}) + \beta \hspace{2pt} \max_{a'} q(s', a')$ \\ 
                    \ \ \ \ \ \  $- q(s, a)$ \\
                $q(s, a) \leftarrow q(s, a) + \alpha \times  \mathrm{err}$ \\
                $s \leftarrow s'$
            }

            $Q(\cdot,\cdot,\tilde{s}) \leftarrow q(\cdot,\cdot)$ 
        }

        \vspace{5pt}
        \For {$\tilde{s} = 0, 1, \cdots, S - 1$}
        {
            $\lambda(\tilde{s}) \leftarrow \lambda(\tilde{s}) + \gamma  \times  (Q(\tilde{s}, 1, \tilde{s}) - Q(\tilde{s}, 0, \tilde{s}))$
        }

        \vspace{5pt}
        \If {$\max_{\tilde{s}}|Q(\tilde{s}, 1, \tilde{s}) - Q(\tilde{s}, 0, \tilde{s})| < \delta$}{
            $\mathrm{break}$
        }
    }

    \vspace{10pt}
    \Return{Indices, $\lambda(s),$ $s \in \mathcal{S}$} \;
    \caption{Index Learning Algorithm}
    \label{algo:II}
\end{algorithm}
Learning of index $\lambda(\tilde{s})$ involves two loops in our Algorithm~\ref{algo:II}. 
In the inner loop, assuming subsidy $\lambda(\tilde{s})$ is fixed for each threshold state $\tilde{s} \in \mathcal{S},$  Q-learning algorithm is performed  and stored in $Q(\cdot, \cdot, \tilde{s}).$  In the outer loop index $\lambda_n(\tilde{s})$ is updated as follows.
\begin{eqnarray*}
    \lambda_{n+1}(\tilde{s}) = \lambda_n(\tilde{s}) + \gamma (Q(\tilde{s}, 1, \tilde{s}) - Q(\tilde{s}, 0, \tilde{s})). 
\end{eqnarray*}
Here, $\gamma$ is constant step size.
This algorithm can be viewed as two timescale algorithm in which subsidy $\lambda_n(\tilde{s})$ is updated on slower timescale and the Q-learning algorithm is performed on faster timescale.  Here we consider constant stepsizes in both Q-learning and index-learning. for given threshold state $\tilde{s} \in \mathcal{S},$ we have  
\begin{eqnarray*}
    Q_{n+1}(s, a, \tilde{s}) =  Q_{n}(s, a, \tilde{s})  + \alpha \left[r(s,a) + \lambda_n(\tilde{s}) (1-a)+ 
    \right. \\ \left. 
    \beta \max_{a^{'}} Q_{n}(s, a^{'}, \tilde{s}) - Q_{n}(s, a, \tilde{s})
    \right]
\end{eqnarray*} 
\begin{eqnarray*}
    \lambda_{n+1}(\tilde{s}) = \lambda_n(\tilde{s}) + \gamma  \left[Q_n(\tilde{s}, 1, \tilde{s}) - Q_n(\tilde{s}, 0, \tilde{s})\right]. 
\end{eqnarray*}
The iterative  algorithms are governed by two step-sizes ($\alpha$ and $\gamma$), where $\alpha$ and $\gamma$ are used in Q-learning and index learning, respectively. We terminate the algorithm in either of the following conditions: either the number of iterations exceeds the pre-defined maximum number of iterations, or the maximum difference, i.e., $\max_{\tilde{s}}|Q_n(\tilde{s}, 1, \tilde{s}) - Q_n(\tilde{s}, 0, \tilde{s})|$ becomes smaller than a pre-defined threshold. 

The analysis of index learning with two-timescale is provided in Appendix.

\section{Index Learning  using DQN}
\label{sec:index-learning-DQN}
We now study DQN algorithm for index learning problem in restless bandits. This is two timescale algorithm with constant stepsize. On faster timescale, we run DQN algorithm and slower timescale, we update the index learning algorithm.  The learning of index $\lambda(\tilde{s})$ for all $\tilde{s} \in \mathcal{S}.$ We say that $\lambda(\tilde{s})$ is the index in SARB, implies that $\tilde{s}$ is a threshold state such that  $Q^{\lambda(\tilde{s})}(\tilde{s},a = 1, \tilde{s}) - Q^{\lambda(\tilde{s})}(\tilde{s},a = 0, \tilde{s}) = 0.$  In DQN algorithm, $Q^{\lambda(\tilde{s})}(\tilde{s},a = 1, \tilde{s}, \theta) $ is an non-linear approximation to $Q$ state-action value function using deep learning neural networks. 

For given $\lambda(\tilde{s}),$ we run DQN algorithm similar to preceding section: use of experience replay with buffer, target  Q-network with weights $\overline{\theta}$ of neural network which are updated periodically  and online Q network where $\theta$ are updated with sampled mini batch  data.  The target value estimate under fixed parameter $\overline{\theta}$ is $y^{\lambda(\tilde{s})}(s,a,\overline{\theta})=r(s,a) +(1-a)\lambda(\tilde{s}) + \beta \max_{a^{\prime}} Q^{\lambda(\tilde{s})}(s^{\prime},a^{\prime},\tilde{s},\overline{\theta})$ for state-action pair $(s,a) \in \mathcal{S}\times \mathcal{A}.$ The squared error loss is $l(s,a,\theta,\tilde{s},\lambda(\tilde{s})),$
\begin{eqnarray*}
    l(s,a,\theta,\tilde{s},\lambda(\tilde{s})) = \frac{1}{2}\left[Q^{\lambda(\tilde{s})}(s,a,\tilde{s},\theta) - y^{\lambda(\tilde{s})}(s,a,\overline{\theta}) \right]^2.
\end{eqnarray*}

The gradient based learning algorithm for $\theta$ is as follows.
\begin{eqnarray}
   \theta_{n+1} = \theta_n - \alpha \frac{\partial l(s,a,\theta,\tilde{s},\lambda(\tilde{s}))}{\partial \theta} \bigg\vert_{\theta= \theta_n}, 
   \label{Eqn:update-rule-theta-2}
\end{eqnarray}
where
\begin{eqnarray*}
 \frac{\partial l(s,a,\theta,\tilde{s},\lambda(\tilde{s}))}{\partial \theta} 
 = \left[ Q^{\lambda(\tilde{s})}(s,a,\tilde{s},\theta) - y^{\lambda(\tilde{s})}(s,a,\overline{\theta}) \right] \times \\
 \frac{\partial Q^{\lambda(\tilde{s})}(s,a,\tilde{s},\theta)}{\partial \theta}.   
\end{eqnarray*}

In every mini-batch  of experiences, $\theta_n$ is used to update online learning in  Eqn.~\eqref{Eqn:update-rule-theta-2}, after mini-batch data, we set $\theta = \theta_n.$
Then we update $\overline{\theta}$ according following rule after every minibatch. 
\begin{eqnarray}
   \overline{\theta} \leftarrow \tau \overline{\theta} + (1-\tau) \theta. 
   \label{Eqn:avg-theta}
\end{eqnarray}
Here $\tau \in (0,1).$ We perform update of $\theta_n$ and $\overline{\theta}$ using Eqn~\eqref{Eqn:update-rule-theta-2} and Eqn.~\eqref{Eqn:avg-theta} for iteration until  $T_{\max}.$  
Note that $\overline{\theta}$ is function of subsidy $\lambda(\tilde{s}).$ This provides us $Q^{\lambda(\tilde{s})}(s,a,\tilde{s},\overline{\theta})$ for given $\tilde{s} \in \mathcal{S}$ and for all $(s,a) \in \mathcal{S}\times \mathcal{A}.$  We further  run these algorithms for all $\tilde{s} \in \mathcal{S}$ in order to obtain $Q^{\lambda(\tilde{s})}(s,a,\tilde{s},\overline{\theta})$  for all $\tilde{s} \in \mathcal{S}$ and  for all $(s,a) \in \mathcal{S}\times \mathcal{A}.$  


In slow timescale algorithm, we update $\lambda(\tilde{s})$ for all $\tilde{s} \in \mathcal{S}$ according to following scheme. 
\begin{eqnarray*}
     \lambda(\tilde{s}) \rightarrow \lambda(\tilde{s}) + \gamma  \left[Q^{\lambda(\tilde{s})}(\tilde{s},a=1,\tilde{s},\overline{\theta}) - Q^{\lambda(\tilde{s})}(\tilde{s},a=0,\tilde{s},\overline{\theta}) \right]. 
\end{eqnarray*}
In our two-timescale analysis, we assume that $0<\gamma<<\alpha<1.$ 
We run two timescale algorithms until either of the following conditions: 1)  the number of iterations exceeds the pre-defined maximum number of iterations, or 2) the maximum difference, i.e., $\max_{\hat{s}}\vert Q^{\lambda(\tilde{s})}(\tilde{s},a=1,\tilde{s},\overline{\theta}) - Q^{\lambda(\tilde{s})}(\tilde{s},a=0,\tilde{s},\overline{\theta})\vert < \delta.$ where $\delta$ is a pre-defined threshold.

\section{Numerical Examples}
\label{sec:numerical examples}
We first illustrate numerical examples for MDP models using Q-learning with different exploration policies ($\epsilon$-greedy, softmax, $\epsilon$-softmax).  
We next illustrate numerical examples for index learning algorithm with Q-learning. Finally present examples for index learning with DQN and function approximations.  

\subsection{Examples for MDP Models}
We plot error function $\Delta V_t$ for $t \in 1, 2, \cdots, T_{max}.$
 It is calculated as follows. 
\begin{equation}
 V_t(s) = \max_{a} Q_t(s,a)   
\end{equation}
and 
\begin{eqnarray}
 \Delta V_t := \sqrt{\frac{1}{|\mathcal{S}|}\sum_{s \in \mathcal{S}}(V_t(s) - V^*(s))^2}   
\end{eqnarray}
Here, $V^*(s)$ the optimal state value function for $s \in \mathcal{S}$ with known model.  When Q-learning algorithm learns optimal Q-function, that is, $\Delta V_t \rightarrow 0$ as $t \rightarrow \infty.$

%


\subsubsection{Example of one-step random walk with $K = 25$}

\begin{figure*}
  \begin{center}
    \begin{tabular}{cccc}
     \includegraphics[scale=0.20]{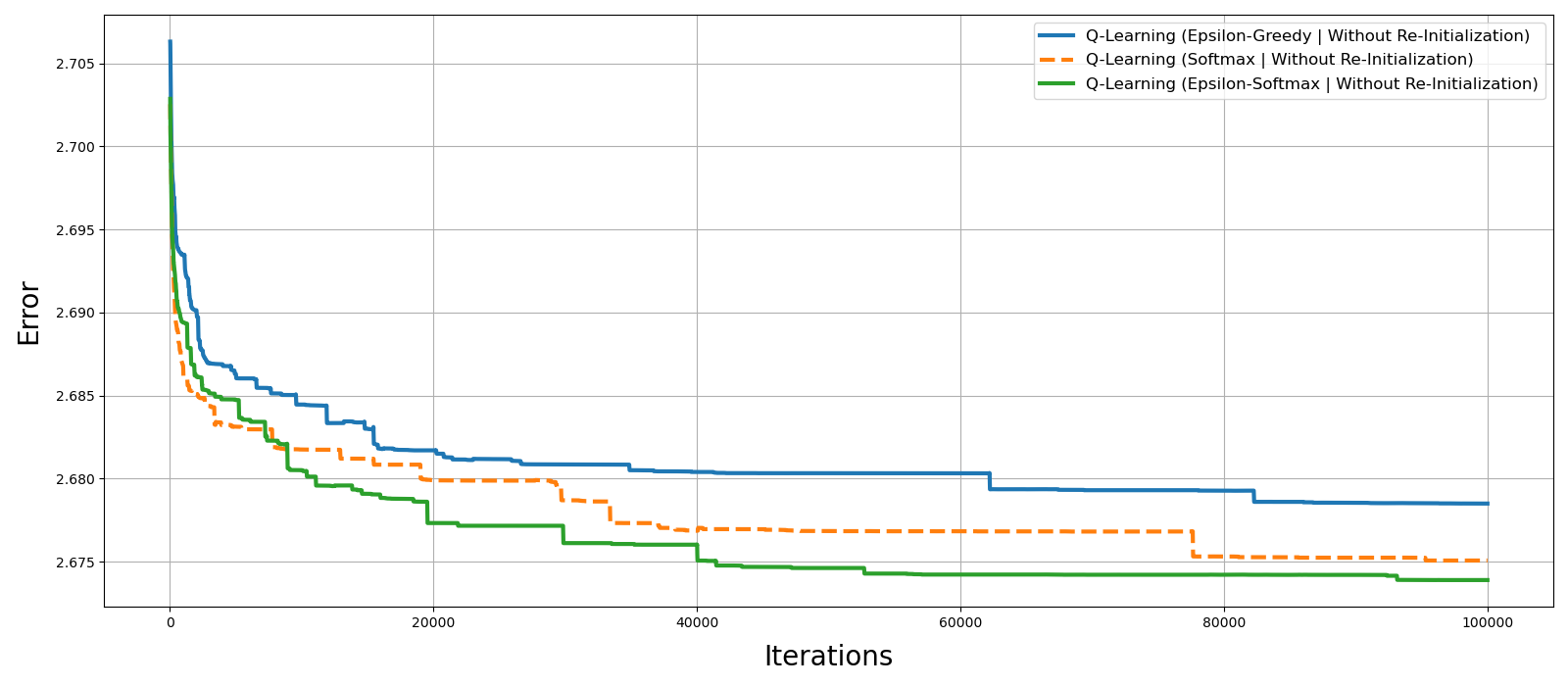}
      & 
      \includegraphics[scale=0.20]{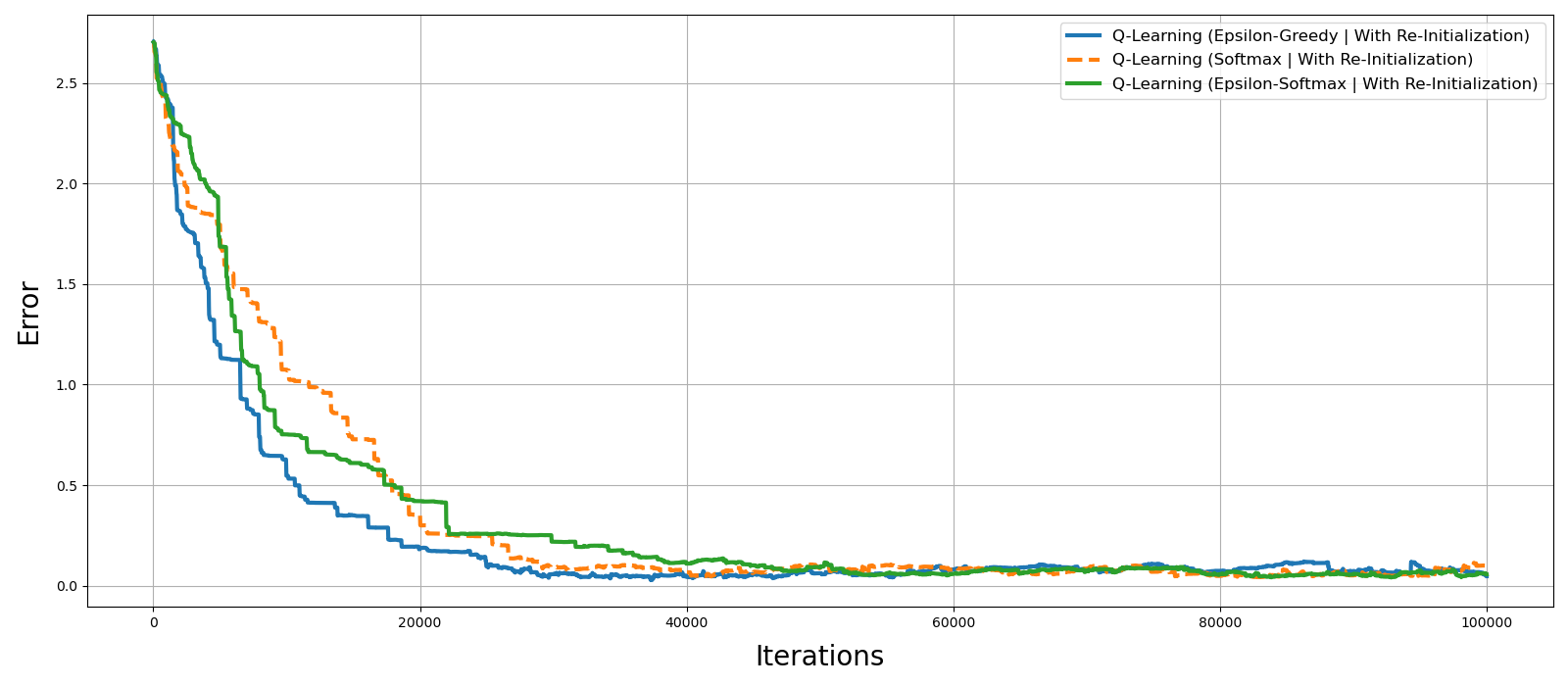}   \\
      a)  error vs iteration without reintialization  & b) error vs iteration with re-initialization 
     \end{tabular}
  \end{center}
  \caption{Q-learning: Example with one step random walk and number of states $K=25$ without and with re-initialization}
  \label{fig:QLearning_05_01_0102-25}
\end{figure*}
We assume that the probability matrix is same for both the actions $a = 1$ and $a = 0$. Rewards for passive action $r(k,0) = 0$ and active action $r(k,1) = 0.95^k$ for all $k \in \mathcal{S}$. This is also applicable in wireless communication systems. 

From Fig.~\ref{fig:QLearning_05_01_0102-25}-a, we can observe that  Q-Learning such as: Q-Learning with $\epsilon$-greedy policy, Q-Learning with softmax policy, and Q-Learning with $\epsilon$-softmax policy are not able to learn the true state values. States which Q value function is not updated, they are having value $0$ and hence error is not converging to $0$ for all Q-learning algorithms.  Figure shows that the error in the state values, i.e., $\Delta V_t$ only falls by a small level, and stabilizes afterwards.
We used parameter values $\beta = 0.9$, $\alpha = 0.2$, $T_{\max}= 100000$ and exploration factor  $\epsilon = 0.4.$

\subsubsection{Example of one-step random walk with $K = 25$ and re-initialization}

We use the same model as the previous one, the only difference being the re-initialization of the current state. The re-initialization of the current state is done in a random manner (reset the state  to random value of state after $50$ iterations to further allow exploration). This is a similar to episodic variant of Q learning. 

From Fig.~\ref{fig:QLearning_05_01_0102-25}-b, we can observe that  Q-Learning with $\epsilon$-greedy policy, Q-Learning with softmax policy, and Q-Learning with $\epsilon$-softmax policy are able to learn the true state values. Fig.~\ref{fig:QLearning_05_01_0102}-b shows that the error in the state values, i.e., $\Delta V_t$ falls steeply in the first $20000$ iterations. In these simulations, after every $50$ iterations, the current state is re-initialized. The rest of the hyper-parameter configuration is kept the same as before.

\subsection{Examples for index learning in SARB using Q-learning }
We now present numerical examples for index learning using Q-learning approach. 
We define error function $\Delta \lambda_k$  as follows
\begin{eqnarray}
 \Delta \lambda_k := \max_{\tilde{s} \in \mathcal{S}} |Q_k(\tilde{s}, 1, \tilde{s}, \lambda_k ) - Q_k(\tilde{s}, 0, \tilde{s}, \lambda_k)|
\end{eqnarray}
We plot $\Delta \lambda_k$ as function of $k$ iteration. We study  examples as discussed in preceding section for index learning. 
\begin{figure*}
  \begin{center}
    \begin{tabular}{cccc}
     \includegraphics[scale=0.20]{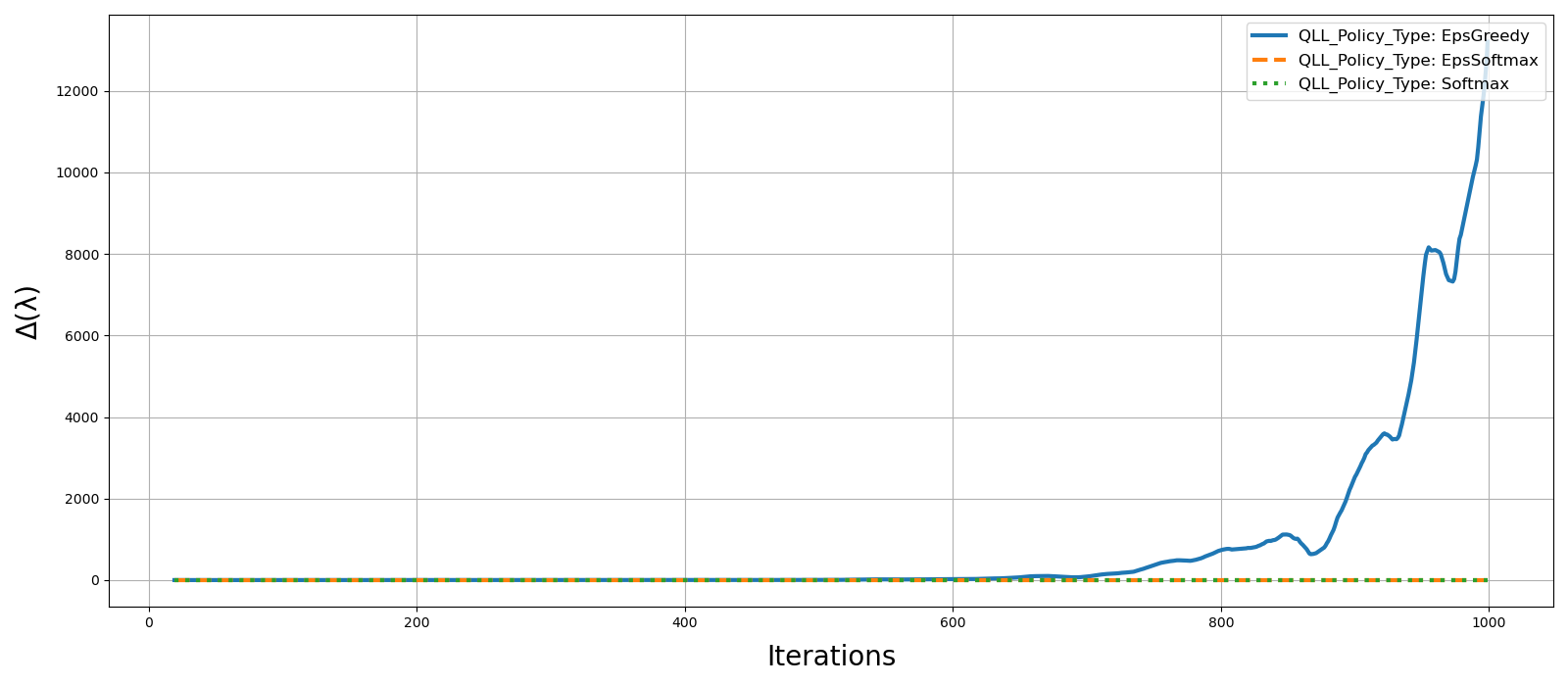}
      & 
      \includegraphics[scale=0.20]{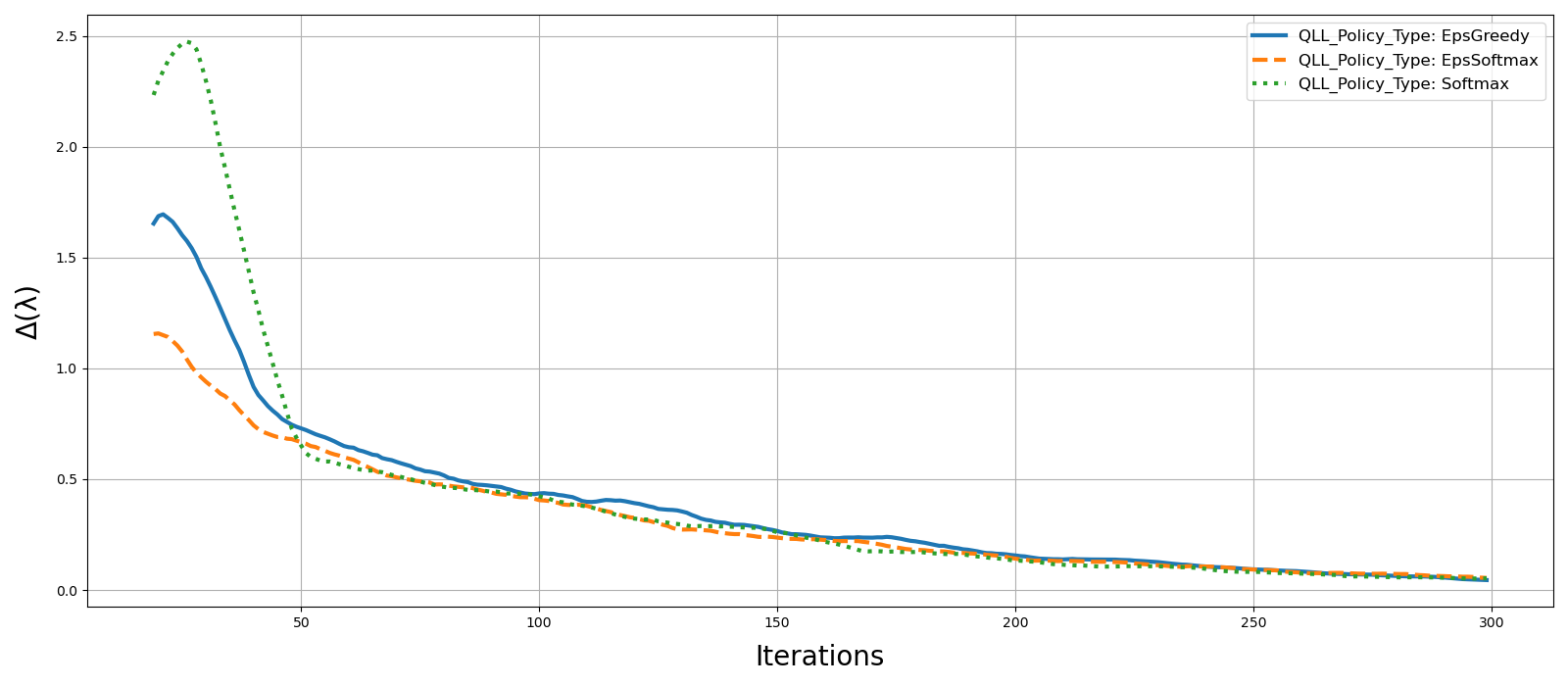}   \\
      a)  error vs iteration & b) error vs iteration with re-initialization
     \end{tabular}
  \end{center}
  \caption{Index learning using Q learning: Example with  One step random walk with $K=25$  }
  \label{fig:QLLNR_05_0102}
\end{figure*} 

\subsubsection{Example of one-step random walk with $K = 25$}
We  consider same examples as discussed in the preceding section for $K=25$ that is, one step random walk. 

From Fig.~\ref{fig:QLLNR_05_0102}-a, it is clearly evident that Q-$\lambda$ learning with all the three policies: $\epsilon$-greedy, softmax, and $\epsilon$-softmax, is unable to find the true indices. The rest of the hyper-parameter configurations are as follows: $\beta = 0.9$, $\epsilon = 0.4$, $K_{\max} = 1000$, $T_{\max} = 5000$, $\alpha = 0.05$, $\gamma = 0.01$, $\delta = 0.005.$
\subsubsection{Example of one-step random walk with $K = 25$ and re-initialization}
We consider re-initialization for one step random walk, and this leads to faster convergence of error to  $0$ which was not possible in earlier example $K=25.$ 
The re-initialization is done as follows: 
\begin{itemize}
    \item $N(s, a)$ = Number of times $Q(s, a)$ is updated (initialized with all 2's to avoid the division by 0 issue)
    \item $N_{state}(s) = \sum_{a \in \mathcal{A}} N(s, a)$
    \item $Inv\_N(s) = \frac{1}{N_{state}(s)}$
    \item $Prob(s) = \frac{Inv\_N(s)}{\sum_{s \in \mathcal{S}} Inv\_N(s)}$
    \item We reset the current state in accordance with $Prob(s)$
\end{itemize}
Re-initialization allows exploration of all state-action pair which did not happen in earlier example.

Fig.~\ref{fig:QLLNR_05_0102}-b shows that the error in the indices, i.e., $\Delta \lambda_k$ falls steeply in the first $150$ iterations for index learning with all the three policies: $\epsilon$-greedy, softmax, and $\epsilon$-softmax. Furthermore, from Fig., we can observe that index learning algorithm with all the three policies is able to learn the true indices. The hyper-parameter configurations are as follows: $\beta = 0.9$, $\epsilon = 0.4$, $K_{\max} = 300$, $T_{\max} = 5000$, $\alpha = 0.05$, $\gamma = 0.01$, $\delta = 0.005.$

\subsection{Examples for index learning using DQN}

\subsubsection{Example of one-step random walk with $K = 5$ and re-initialization}
We consider one step random walk with $K=5$ The probability matrix is same for both the actions $a = 1$ and $a = 0$. Rewards for passive action $r(k,0) = 0$ and active action $r(k,1) = 0.9^k$ for all $k \in \mathcal{S}$. This is also applicable in wireless communication systems. The probability matrices are given for $K = 5$ states.
\begin{eqnarray*}
  P_0 &=&  \begin{bmatrix}
    3/10 & 7/10 & 0 & 0 & 0 \\
    1/10 & 2/10 & 7/10 & 0 & 0 \\
    0 & 1/10 & 2/10 & 7/10 & 0 \\
    0 & 0 & 1/10 & 2/10 & 7/10 \\
    0 & 0 & 0 & 3/10 & 7/10
\end{bmatrix}, \\
 P_1 &=& P_0.
\end{eqnarray*}

\begin{figure}
  \begin{center}
    \begin{tabular}{cccc}
     \includegraphics[scale=0.23]{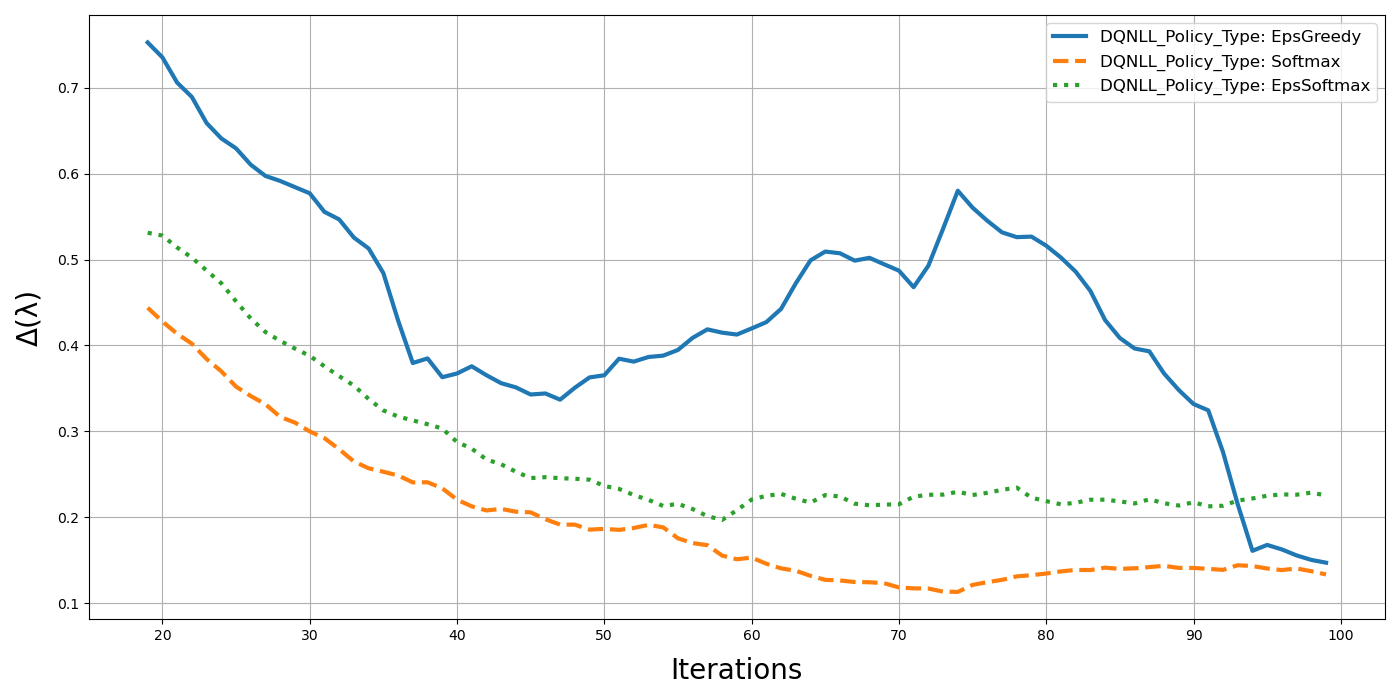}
     \end{tabular}
  \end{center}
  \caption{index learning with DQN algorithm Example: one step random walk $K=5$ with re-intialization}
  \label{fig:DQNLL_04_0102}
\end{figure} 

Fig.~\ref{fig:DQNLL_04_0102} shows that the error in the indices, i.e., $\Delta \lambda_k$ falls steeply in the first 50 iterations for DQN-$\lambda$ learning with softmax and $\epsilon$-softmax policies. However, the error fluctuates for DQN-$\lambda$ learning with $\epsilon$-greedy policy. 
The rest of the hyper-parameter configurations are as follows: $memory\_size = 10000$, $mb\_size = 64$, $\alpha_1 = 0.01$, $\alpha_2 = 0.05$, $\gamma = 0.9$, $\tau = 0.001$, $\delta = 0.001$, $T_{max} = 500$, $K_{max} = 100$, $c = 1$. $\epsilon$ is initialized as $0.1$, with a decay rate of $0.99$. It is decayed after every $20$ time-steps, until it reaches $0.01$. For $oq\_network$ and $tq\_network$, number of hidden units in each hidden layer = $32$, and number of hidden layers = $2$.

\subsection{Q-$\lambda$ Learning + Linear Function Approximation}

For larger state spaces, we cannot use index learning with $Q$ learning, since they inherently employ a tabular approach to update the action values for each state-action pair, which quickly becomes computationally infeasible as the number of states increases. Additionally, it is not guaranteed anymore that each state-action pair will be visited a large number of times, which is one of the key requirements for Q-$\lambda$ learning to function appropriately. Both of these disadvantages can be tackled with the help of function approximation. 

In the previous section, we have seen DQN-$\lambda$ learning, which employs non-linear function approximation, but as we observed, it gives poor performance. Additionally, we observed that the computation time for DQN-$\lambda$ learning is huge, which can be easily justified by observing the computation time for DQN. Consider the "Example of one-step random walk with  $K = 5$", for which DQN roughly takes $20$ mins for $20000$ iterations, i.e. $0.001$ min for every iteration. For the same example only, DQN-$\lambda$ learning is run for $K_{Iters} = 100$ and $T_{Iters} = 500$. where it requires $T_{Iters}$ for every $\hat{s} \in \mathcal{S}$. So, the computation time for DQN in index learning would be roughly $0.001 * 500 * 5 * 100 = 250$ mins. 

To tackle these issues,  we consider index learning with linear function approximation, so that the algorithm can perform suitably well in larger state-spaces.

\subsubsection{Example of one-step random walk with $K = 500$ and re-initialization}
\begin{figure*}
  \begin{center}
    \begin{tabular}{cccc}
     \includegraphics[scale=0.23]{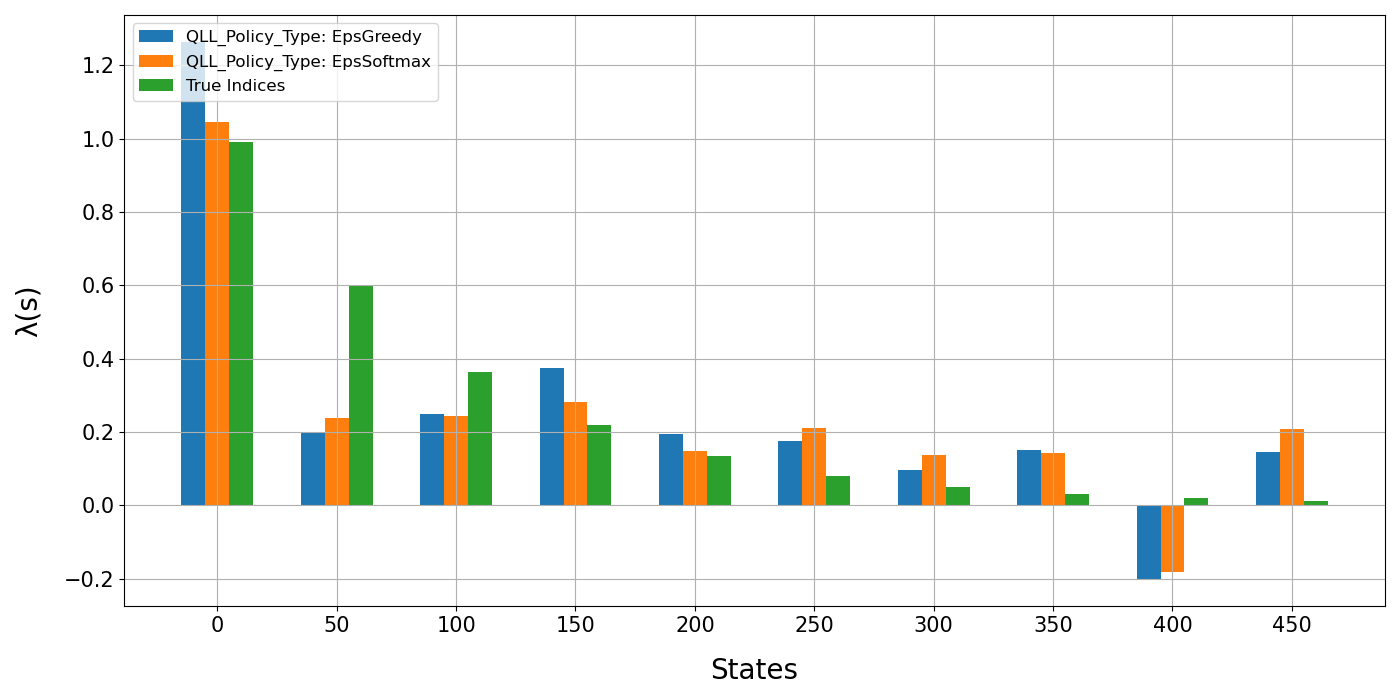}
      & 
      \includegraphics[scale=0.23]{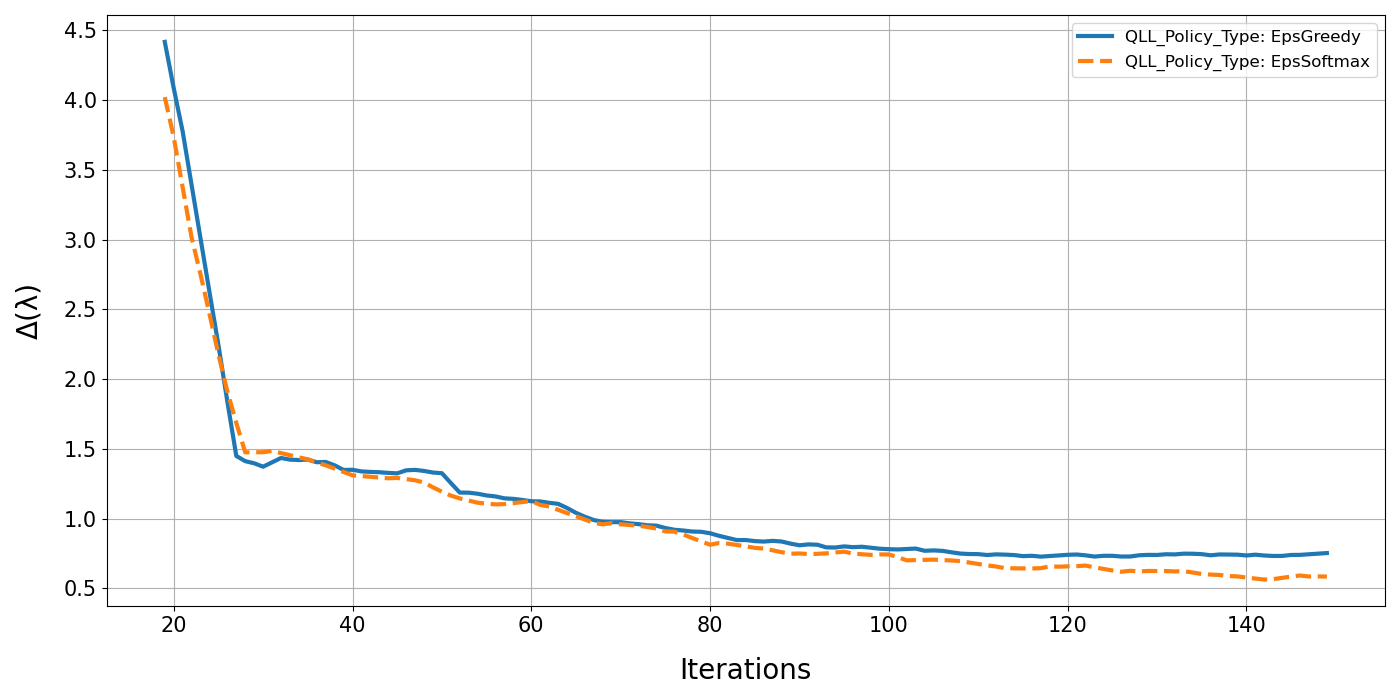}   \\
      a)  index values vs state & b) error vs iteration
     \end{tabular}
  \end{center}
  \caption{Linear function approximation: Example of one-step random walk with $K = 500$ and re-initialization}
  \label{fig:QLLLFA_02_0102}
\end{figure*} 


Fig.~\ref{fig:QLLLFA_02_0102} shows the true indices across the states. Fig.~\ref{fig:QLLLFA_02_0102}-b shows that the error in the indices, i.e., $\Delta \lambda_k$ falls steeply in the first 30 iterations for Q-$\lambda$ learning with $\epsilon$-greedy and $\epsilon$-softmax policies. Furthermore, we can observe that Q-$\lambda$ learning with $\epsilon$-greedy and $\epsilon$-softmax policies is able to learn the true indices, to a partial extent. The rest of the hyper-parameter configurations are as follows: $\beta = 0.9$, $\epsilon = 0.4$, $K_{max} = 150$, $T_{max} = 3000$, $\alpha = 0.05$, $\gamma = 0.01$, $\delta = 0.001$. For linear function approximation, we have employed state aggregation with a group size of $10$ states each, i.e., $50$ groups.

\section{Concluding remarks}
\label{sec:conclusion}

We studied Whittle index learning algorithm with Q-learning, DQN and function approximations. We analyzed constant stepsizes two timescale stochastic approximation algorithms for index learning. We also  studied exploration schemes $\epsilon$-greedy, softmax and $\epsilon$-softmax policies. We observed that the performance of exploration policies is model dependent. We also noted that the DQN based index learning is computationally time expensive, hence linear function approximations such as linear state aggregation is better suited approach. However numerical examples illustrate overall good performance of index learning with Q learning with reasonable state space.

\section{Acknowledgment}
Part of this work was carried  at IIIT Allahabad. 
The work of Rahul Meshram at IIT Madrad is supported by NFI Grant IIT Madras and SERB Project “Weakly Coupled POMDPs and Reinforcement Learning”.

\appendix 
\subsection{Convergence of two-timescales stochastic approximation algorithms}
We discuss convergence of two-timescale SA  with decreasing stepsizes and later discuss constant stepsizes SA. 
Consider two-timescale SA with decreasing stepsizes
\begin{eqnarray}
    x_{n+1} = x_n + \alpha_n [h(x_n,y_n) + M_{n+1}^1] \\ 
     y_{n+1} = y_n + \gamma_n [g(x_n,y_n) + M_{n+1}^2] 
\end{eqnarray}
where $x_n \in \mathbb{R}^d_1,$ $y_n \in \mathbb{R}^d_2,$ and  $h: \mathbb{R}^{d_1 + d_2} \rightarrow \mathbb{R}^{d_1},$  $g: \mathbb{R}^{d_1 + d_2} \rightarrow \mathbb{R}^{d_2}$ are Lipschitz continuous functions. $M_n^{i}$ are martingale noise terms for $i=1,2.$ $\alpha_n$ and $\gamma_n$ are stepsizes, $\alpha_n \rightarrow 0,$ $\gamma_n \rightarrow 0,$ $\sum_{n=1}^{\infty} \alpha_n = \sum_{n=1}^{\infty} \gamma_n = \infty,$ $\frac{\gamma_n}{\alpha_n} \rightarrow 0$ as $n \rightarrow \infty.$ The condition $\frac{\gamma_n}{\alpha_n} \rightarrow 0$ implies that $y_n$ updates on slower timescale, as it can be viewed as stitic, and $x_n$ updates on faster timescale. Using SA analysis, we want $x_n \rightarrow x^*$ and $y_n \rightarrow y^*,$ where $x_n^* \mathbb{R}^{d_1}$ and $y_n^* \mathbb{R}^{d_2}.$ Necessary assumption is boundedness of iterates, $\sup_{n} ||x_n|| < \infty,$ $\sup_{n} ||y_n|| < \infty.$

Convergence of SA iterates require following assumptions.
\begin{enumerate}
    \item $h$ and $g$ are Lipschitz  continuous functions.
    \item $\{M_n^1\}$ and $\{M_n^2\}$ are martingale difference sequences with respect to increasing $\sigma$ fields $\{\mathcal{F}_n\},$ $\mathcal{F}_n = \sigma\left( x_m,y_m, M_m^1,M_m^2, m\leq n\right)$ $n\geq 0$ and it satisfy
    $\expect{||M_{n+1}^{i}||^2~|~ \mathcal{F}_n} \leq K (1+ ||x_n||^2 + ||y_n||^2)$ for $i=1,2,$ $n \geq 0$ $K>0.$
    \item $\{ \alpha_n \},$ $\{\gamma_n\}$ are stepsizes, $\alpha_n > 0,$ $\gamma_n>0,$ $\sum_{n} \alpha_n  = \sum_n \gamma_n = \infty,$ $\sum_n \left( \alpha_n^2 + \gamma_n^2\right) < \infty,$ $\frac{\gamma_n}{\alpha_n} \rightarrow 0$ as $n \rightarrow \infty. $
    \item The function $h_c(x,y) := \frac{h(cx,cy)}{c},$ $c\geq 1$ satisfy $h_c \rightarrow h_{\infty}$ as $c \rightarrow \infty$ uniformly on compacts for some $h_{\infty}.$ The limiting o.d.e.  $\dot{x}(t) = h_{\infty}(x(t),y)$ has unique globally asymptotically stable equilibrium $x^{*}_{\infty}(y): \mathbb{R}^{d_2} \rightarrow \mathbb{R}^{d_1}$ Lipschtiz map, $x^{*}_{\infty}(0) = 0,$ and $\dot{x}(t) = h_{\infty}(x(t),0)$ has origin in $\mathbb{R}^d_1$ as unique globally asymptotically stable equilibrium.
    \item The function $g_c(y) := \frac{g(cx^{*}_{\infty}(y), cy)}{c},$ $c\geq 1$ satisfy $g_c \rightarrow g_{\infty}$ as $c \rightarrow \infty$ uniformly on compacts for some $g_{\infty}.$ 
    The limiting o.d.e. $\dot{y} = g(x^{*}_{\infty}(y(t)),y(t))$ has unique globally asymptotically stable equilibrium.
\end{enumerate} 
In fast-timescale,  SA algorithm tracks the solution of ODE $\dot{x}(t) = h_r(n)(x(t),y(t))$ and $\dot{y}(t) = 0.$ The growth of fast timescale iterate is bounded by slow timescale variable $y_n.$ In slow timescale analysis, for $n$ large enough, the growth of slow timescale variable $y_n$ is bounded. This implies that the growth of fast timescale iterate is bounded.  Then, the scaled version of algorithm tracks the scaled ODE and this in turn tracks the limiting ODE. For detailed analysis, see \cite{Laxminarayanan17}. 

This analysis can be extended to index learning using Q-learning in  a restless bandit problem. The updates in Q-learning are asynchronous, i.e., not all state-action pair updated in each time step. By appropriate choices of stepsizes, one can obtain the convergence of asynchronous Q-learning iterate in two-timescale algorithm. The Q learning update with decreasing step sizes are as follows.
\begin{eqnarray*}
    Q_{n+1}(s, a, \tilde{s}) =  Q_{n}(s, a, \tilde{s})  + \alpha_n \left[r(s,a) + \lambda_n(\tilde{s}) (1-a)+ 
    \right. \\ \left. 
    \beta \max_{a^{'}} Q_{n}(s, a^{'}, \tilde{s}) - Q_{n}(s, a, \tilde{s})
    \right]
\end{eqnarray*} 
\begin{eqnarray*}
    \lambda_{n+1}(\tilde{s}) = \lambda_n(\tilde{s}) + \gamma_n \left[Q_n(\tilde{s}, 1, \tilde{s}) - Q_n(\tilde{s}, 0, \tilde{s})\right]. 
\end{eqnarray*}
Here we mention steps involved in  convergence proof. 
\begin{enumerate}
    \item Limiting ODE for Q-learning algorithm assuming that the subsidy $\lambda_n= \lambda$ is static, is  
    \begin{eqnarray}
        \dot{Q}_{\lambda}(t) = h(Q(t),\lambda), 
    \end{eqnarray}
    where $h(Q(t),\lambda) = F(Q_{\lambda}(t))-Q_{\lambda}(t).$ It has  asymptotic globally stable equilibrium which is unique as $F(Q_{\lambda}(t))$ is contraction mapping. $Q^*_{\lambda}$ is asymptotically global stable equilibrium.
    \item  Limiting o.d.e. for index learning $\lambda_n$ is 
    \begin{eqnarray}
        \dot{\lambda}(t)  = g(\lambda(t),
    \end{eqnarray}
    where $g(\tilde{s},\lambda(t)) = Q^*_{\lambda(t)}(\tilde{s},1)- Q^*_{\lambda(t)}(\tilde{s},0).$
    Further, $Q^*_{\lambda(t)}$ is Lipshitz in $\lambda.$
    \item The growth of noise $M_{n}^{1},$ and $M_{n}^{2}$ can be described in terms of $M_{n}(\tilde{s}),$ it is martingale difference sequence, and $\expect{ ||M_{n+1}(\tilde{s}) ||^2~|~ \mathcal{F}_n} \leq K \left( 1+ ||Q_n(\tilde{s})||^2 + ||\lambda_n(\tilde{s})||^2\right)||.$
    \item With decreasing stepsizes which satisfy following property in index learning: $\alpha_n$ and $\gamma_n$ are stepsizes, $\alpha_n \rightarrow 0,$ $\gamma_n \rightarrow 0,$ $\sum_{n=1}^{\infty} \alpha_n = \sum_{n=1}^{\infty} \gamma_n = \infty,$ $\frac{\gamma_n}{\alpha_n} \rightarrow 0$ as $n \rightarrow \infty.$ 
    \item The scaled ode has $h_{c}(Q,\lambda) =  \frac{h(c Q, c \lambda }{c}).$ As $c \rightarrow \infty,$ $h_c (Q,\lambda) \rightarrow h_{\infty} (Q,\lambda),$ uniformly on compact set  and limiting ODE $\dot{Q}(t) = h_{\infty} (Q(t),\lambda).$ Here $h_{\infty} (Q(t),\lambda) = F_{\infty}(Q_{\lambda}(t),\lambda) - Q_{\lambda}(t).$ Here, $h_{\infty} (Q(s,a),\lambda) = \lambda + \beta \sum_{s^{\prime} \in \mathcal{S} } p_{s,s^{\prime}}^a  
   \max_{a^{\prime}} Q(s^{\prime},a^{\prime}) - Q(s,a).  $  
    Assuming $\lambda(t) =\lambda$  is static, one analyzes the fast timescale. Note that $F_{\infty}$  is a contraction with respect to max-norm $||\cdot ||_{\infty}.$ It has unique asymptotically stable equilibrium $\widehat{Q}^*_{\lambda}.$ It has $\dot{Q}(t) = F_{\infty}(Q(t),0) - Q(t)$  for $\lambda =0,$ and it has origin in $\mathbb{R}^d$ as its unique globally asymptotically stable equilibrium. For large $n$ and large $c >>1,$ the faster timescale iterate are within $\epsilon$ neighborhood of the stable equilibrium of the limiting ode. 
    \item One next consider scaled ODE for slow timescale algorithm. Let $g_c(\lambda) = \frac{g(Q_{c\lambda}, c\lambda)}{c}$ and $g_{\infty}(\lambda)  = \lim_{c \rightarrow \infty} g_{c}(\lambda).$ Define $\widehat{r}_c(s,a) = \frac{r(s,a) + (1-a) c \lambda}{c}.$ 
    Further, $\frac{Q^{(c)(s,a)}}{c} \rightarrow Q^{\infty}_{\lambda}$ as $c \rightarrow \infty$ and 
    $Q^{\infty}_{\lambda}(s,a)= \lambda (1-a) + \beta \sum_{s^{\prime} \in \mathcal{S}} p_{s,s^{\prime}}^a \max_{a^{\prime}} Q(s^{\prime},a^{\prime}).$ For $\lambda >0,$ and $c \rightarrow \infty,$ eventually optimal action is $a= 0$ (not play) for all states and it implies that $Q^{\infty}_{\lambda}(s,a=1) = \beta Q^{\infty}_{\lambda}(s,0) $ and $Q^{\infty}_{\lambda}(s,a=0) = \lambda + \beta Q^{\infty}_{\lambda}(s,0) $. Thus $Q^{\infty}_{\lambda}(s,a=1) - Q^{\infty}_{\lambda}(s,a=0) = -\lambda.$ Similarly for $\lambda < 0, $ $c \rightarrow \infty,$ eventually optimal action is $a= 1$ ( play) for all states, and hence we can  $Q^{\infty}_{\lambda}(s,a=1) - Q^{\infty}_{\lambda}(s,a=0) = -\lambda.$ At $\lambda =0,$ we have $Q^{\infty}_{\lambda}(s,a=1) - Q^{\infty}_{\lambda}(s,a=0) =0.$ Thus $g_{\infty}(\lambda) =-\lambda. $ Then limiting ODE $\dot{\lambda}(t) = -\lambda(t)$ and it has zero as unique globally asymptotically stable equilibrium. From \cite[Theorem 10]{Laxminarayanan17}, slow timescale iterate are bounded and its tracks the limiting ODE and iterate goes to unit ball around origin at the exponential rate. 
\end{enumerate}
We mimic the proof ideas  from \cite{Avrachenkov2022}. When Q-learning updates are asynchronous, the limiting ODE is $\dot{Q}(t) = \Gamma_t h(Q(t), \lambda(t)),$ where $\Gamma_t$ is diagonal matrix with non-negative entries. for each $t \geq 0.$ Entries in matrix defines frequency of update of components. From \cite[Theorem $1$]{Avrachenkov2022}, we get convergence of  $\lambda_k(\tilde{s}) $ to index $\lambda^{*}(\tilde{s})$ a.s. (almost surely).

\subsection{Two-timescale constant stepsizes stochastic approximation algorithms}
Constant stepsizes SA algorithms are as follows
\begin{eqnarray}
    x_{n+1} = x_n + \alpha [h(x_n,y_n) + M_{n+1}^1] \\ 
     y_{n+1} = y_n + \gamma [g(x_n,y_n) + M_{n+1}^2] 
\end{eqnarray}
for $0<\gamma<<\alpha.$ $h,g$ are Lipshcitz. $\expect{||M_{n+1}^{i}||^2~|~ \mathcal{F}_n} \leq K (1+ ||x_n||^2 + ||y_n||^2)$ for $i=1,2,$ $n \geq 0$ $K>0.$

The function $h_c(x,y) := \frac{h(cx,cy)}{c},$ $c\geq 1$ satisfy $h_c \rightarrow h_{\infty}$ as $c \rightarrow \infty$ uniformly on compacts for some $h_{\infty}.$ The limiting o.d.e.  $\dot{x}(t) = h_{\infty}(x(t),y)$ has unique globally asymptotically stable equilibrium $x^{*}_{\infty}(y): \mathbb{R}^{d_2} \rightarrow \mathbb{R}^{d_1}$ Lipschtiz map, $x^{*}_{\infty}(0) = 0,$ and $\dot{x}(t) = h_{\infty}(x(t),0)$ has origin in $\mathbb{R}^d_1$ as unique globally asymptotically stable equilibrium.
 The function $g_c(y) := \frac{g(cx^{*}_{\infty}(y), cy)}{c},$ $c\geq 1$ satisfy $g_c \rightarrow g_{\infty}$ as $c \rightarrow \infty$ uniformly on compacts for some $g_{\infty}.$ 
    The limiting o.d.e. $\dot{y} = g(x^{*}_{\infty}(y(t)),y(t))$ has unique globally asymptotically stable equilibrium. Following the analysis from \cite[Chapter $8$]{Borkar08}, and previous discussion of decreasing stepsize, one can conclude that 
    \begin{eqnarray*}
      \lim\sup_{n \rightarrow \infty} \expect{||x_n-x_{\infty}^{*}(y^*) || + ||y_n-y^*||^2} = O(\alpha) + O\left(\frac{\gamma}{\alpha}\right).
    \end{eqnarray*} 
The stability analysis of SA can be extended for constant stepsize and condition is here weaker than decreasing stepsize. We can state the folloiwng.     
For sufficiently small value of $\alpha,$ analyzing fast timescale, $\sup_n \expect{ || x_n(y)||^2} < \infty$ for all $y.$ In slow timescale analysis, $\sup_n \expect{ || y_n||^2} < \infty.$  

This can be applied to Q-learning discussed in preceding section, we can have 
$\sup_{n} \expect{||Q_n(\lambda)||^2} < \infty$  for all $\lambda$ and 
$\sup_{n} \expect{||\lambda_n||^2} < \infty.$ Further, with constant stepsize, we have 
\begin{eqnarray*}
 \limsup_{n\rightarrow \infty} \expect{||Q_n(\lambda) - Q^{\infty}_{\lambda}||} =   O(\alpha) + O\left(\frac{\gamma}{\alpha}\right).  
\end{eqnarray*}
and 
\begin{eqnarray*}
 \limsup_{n\rightarrow \infty} \expect{||\lambda_n - \lambda^*||} =   O(\alpha) + O\left(\frac{\gamma}{\alpha}\right).  
\end{eqnarray*}
Here, $Q^{\infty}_{\lambda}$ is solution of limiting ode. $h_{\infty}$ and $\lambda^*$ is solution of limiting ode $g_{\infty}.$ It has origin as unique asymptotically stable equilibrium. 
Hence 
\begin{eqnarray*}
 \limsup_{n\rightarrow \infty} \expect{||Q_n - Q^{\infty}_{\lambda^*}|| + ||\lambda_n - \lambda^*||} = \\  O(\alpha) + O\left(\frac{\gamma}{\alpha}\right).  
\end{eqnarray*}

\newpage 
\subsection{Comparison of Q learning and DQN in terms of computation time}

In the below table, we can find the comparison of Q-Learning with Deep Q-Network (DQN). The compute time is in minutes. 
\vspace{10pt}

\begin{tabular}{|p{1.8cm}||p{1.4cm}|p{1.4cm}|p{1.4cm}|}
\hline
& \#Iterations & Compute Time & Error\\
\hline

\multicolumn{4}{|c|}{Example with circular dynamics} \\
\hline
QL (EG)  & 30000 & 0.04  & 0.064\\
QL (SO)  & 30000 & 0.04  & 0.044\\
QL (ES)  & 30000 & 0.04  & 0.046\\
DQN (EG) & 20000 & 20.53 & 0.118\\
DQN (SO) & 20000 & 21.29 & 0.425\\
DQN (ES) & 20000 & 21.20 & 0.088\\
\hline

\multicolumn{4}{|c|}{Example with no structure on transition model} \\
\hline
QL (EG)  & 30000 & 0.04  & 0.041\\
QL (SO)  & 30000 & 0.05  & 0.030\\
QL (ES)  & 30000 & 0.05  & 0.069\\
DQN (EG) & 20000 & 23.31 & 0.061\\
DQN (SO) & 20000 & 23.73 & 0.192\\
DQN (ES) & 20000 & 23.88 & 0.043\\
\hline

\multicolumn{4}{|c|}{Example with restart model} \\
\hline
QL (EG)  & 30000 & 0.04  & 0.067\\
QL (SO)  & 30000 & 0.04  & 1.353\\
QL (ES)  & 30000 & 0.05  & 0.097\\
DQN (EG) & 20000 & 23.31 & 0.020\\
DQN (SO) & 20000 & 23.46 & 0.390\\
DQN (ES) & 20000 & 23.00 & 0.024\\
\hline

\multicolumn{4}{|c|}{Example of one-step random walk with $K = 5$} \\
\hline
QL (EG)  & 100000 & 0.11 & 0.050\\
QL (SO)  & 100000 & 0.30 & 0.692\\
QL (ES)  & 100000 & 0.37 & 0.047\\
DQN (EG) & 20000 & 20.40 & 0.052\\
DQN (SO) & 20000 & 21.07 & 0.314\\
DQN (ES) & 20000 & 20.69 & 0.219\\
\hline

\end{tabular}

\begin{tabular}{|p{1.8cm}||p{1.4cm}|p{1.4cm}|p{1.4cm}|}
\hline
& \#Iterations & Compute Time & Error\\
\hline

\multicolumn{4}{|c|}{Example of one-step random walk with $K = 25$} \\
\hline
QL (EG)  & 100000 & 0.12 & 2.679\\
QL (SO)  & 100000 & 0.24 & 2.675\\
QL (ES)  & 100000 & 0.46 & 2.674\\
DQN (EG) & 20000 & 24.10 & 2.086\\
DQN (SO) & 20000 & 23.78 & 2.071\\
DQN (ES) & 20000 & 24.34 & 2.045\\
\hline

\multicolumn{4}{|c|}{Example of one-step random walk with $K = 25$} \\
\multicolumn{4}{|c|}{and re-initialization} \\
\hline
QL (EG)  & 100000 & 0.12 & 0.047\\
QL (SO)  & 100000 & 0.44 & 0.099\\
QL (ES)  & 100000 & 0.49 & 0.056\\
DQN (EG) & 10000 & 12.76 & 0.067\\
DQN (SO) & 10000 & 12.97 & 0.484\\
DQN (ES) & 10000 & 12.89 & 0.045\\
\hline

\end{tabular}

\newpage 
\subsection{ Comparison of index learning using  Q learning and DQN in terms of computation time}
\vspace{10pt}
In the below table, we can find the comparison of Q-$\lambda$ Learning with DQN-$\lambda$ Learning. The compute time is in minutes. 
\vspace{10pt}

\begin{tabular}{|p{2cm}||p{1cm}|p{1cm}|p{1.2cm}|p{1cm}|}
\hline
& $K_{Iters}$ & $T_{Iters}$ & Compute Time & Error\\
\hline

\multicolumn{5}{|c|}{Example with circular dynamics} \\
\hline
QLL (EG)   & 1000 & 5000 & 14.45 & 0.032\\
QLL (SO)   & 1000 & 5000 & 11.58 & 0.044\\
QLL (ES)   & 1000 & 5000 & 14.18 & 0.048\\
\hline

\end{tabular}

\begin{tabular}{|p{2cm}||p{1cm}|p{1cm}|p{1.2cm}|p{1cm}|}
\hline
& $K_{Iters}$ & $T_{Iters}$ & Compute Time & Error\\
\hline

\multicolumn{5}{|c|}{Example with no structure on transition model} \\
\hline
QLL (EG)   & 1000 & 5000 & 420.73 & 0.016\\
QLL (ES)   & 1000 & 5000 & 19.80 & 0.003\\
\hline

\multicolumn{5}{|c|}{Example with restart model} \\
\hline
QLL (EG)   & 1000 & 5000 & 27.00 & 0.004\\
QLL (ES)   & 1000 & 5000 & 37.71 & 0.005\\
\hline

\multicolumn{5}{|c|}{Example of one-step random walk with $K = 5$} \\
\hline
QLL (EG)   & 1000 & 5000 & 23.80 & 0.034\\
QLL (SO)   & 1000 & 5000 & 34.87 & 0.021\\
QLL (ES)   & 1000 & 5000 & 32.48 & 0.007\\
\hline

\multicolumn{5}{|c|}{Example of one-step random walk with $K = 5$} \\
\multicolumn{5}{|c|}{and re-initialization} \\
\hline
DQNLL (EG) & 100 & 500 & 315.31 & 0.131\\
DQNLL (SO) & 100 & 500 & 321.51 & 0.079\\
DQNLL (ES) & 100 & 500 & 319.90 & 0.172\\
\hline

\end{tabular}

\begin{tabular}{|p{2cm}||p{1cm}|p{1cm}|p{1.2cm}|p{1cm}|}
\hline
& $K_{Iters}$ & $T_{Iters}$ & Compute Time & Error\\
\hline

\multicolumn{5}{|c|}{Example of one-step random walk with $K = 25$} \\
\hline
QLL (EG)   & 1000 & 5000 & 698.17 & 19797.5\\
QLL (SO)   & 1000 & 5000 & 787.31 & 0.864\\
QLL (ES)   & 1000 & 5000 & 769.60 & 0.864\\
\hline

\multicolumn{5}{|c|}{Example of one-step random walk with $K = 25$} \\
\multicolumn{5}{|c|}{and re-initialization} \\
\hline
QLL (EG)   & 300 & 5000 & 66.56 & 0.040\\
QLL (SO)   & 300 & 5000 & 100.31 & 0.057\\
QLL (ES)   & 300 & 5000 & 93.23 & 0.039\\
\hline

\end{tabular}

\newpage 
\subsection{Additional Numerical Examples: Q learning} 
In this section, we present additional numerical examples for Q-learning with different models. 

\subsubsection{Example with circular dynamics}
This example is studied in \cite{Avrachenkov2022, Fu2019}, model has four states and the dynamics are circular: when an arm is passive ($a = 0$), resp. active ($a = 1$), the state evolves according to the transition probability matrices. 
\begin{eqnarray*} 
P^0 = 
\begin{bmatrix}
1/2 & 0 & 0 & 1/2 \\
1/2 & 1/2 & 0 & 0 \\
0 & 1/2 & 1/2 & 0 \\
0 & 0 & 1/2 & 1/2
\end{bmatrix},
\end{eqnarray*}
\begin{eqnarray*} 
 P^1 = {P^0}^T.
\end{eqnarray*}
The reward matrix is given as follows.
\begin{eqnarray*} 
R = 
\begin{bmatrix}
-1 & -1 \\
0 & 0 \\
0 & 0  \\
1 & 1
\end{bmatrix}.
\end{eqnarray*}

\begin{figure}
  \begin{center}
    \begin{tabular}{cc}
     \includegraphics[scale=0.23]{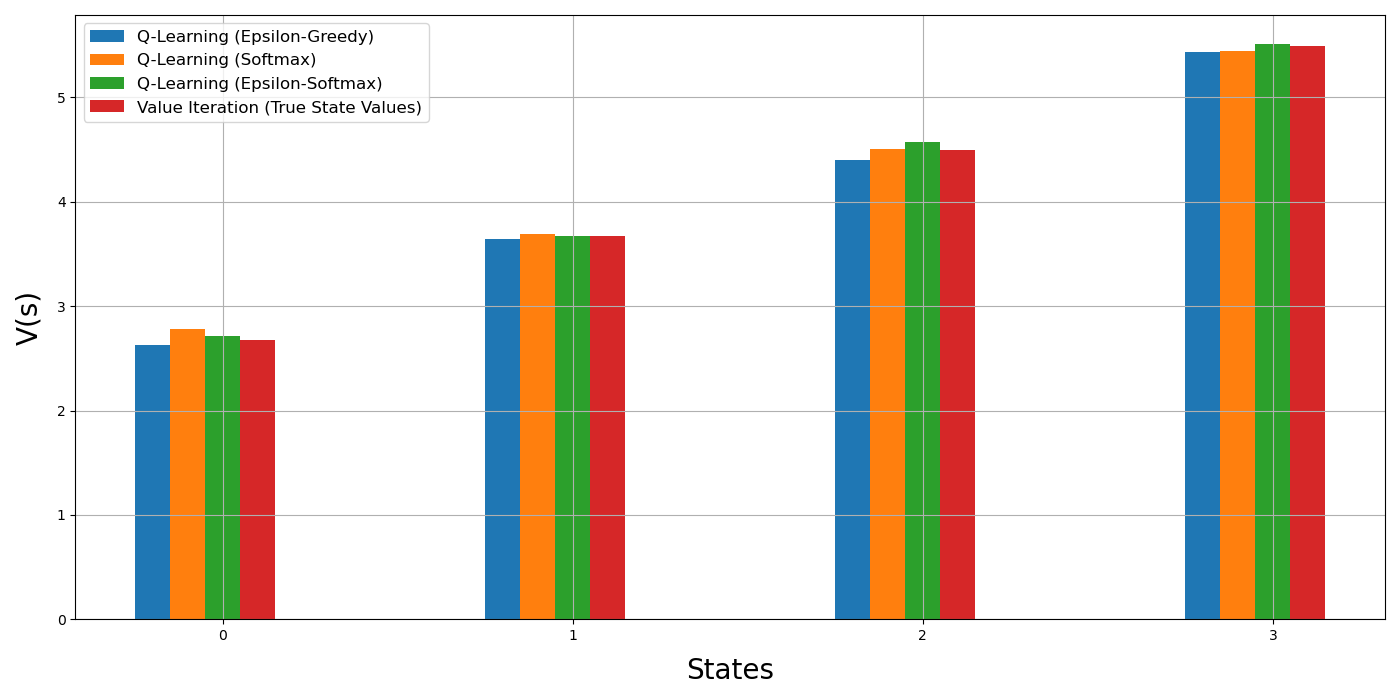}
     \end{tabular}
  \end{center}
  \caption{Q-learning: Example with circular dynamic model }
  \label{fig:QLearning_01_0102a}
\end{figure}

\begin{figure}
  \begin{center}
    \begin{tabular}{cc}
      \includegraphics[scale=0.23]{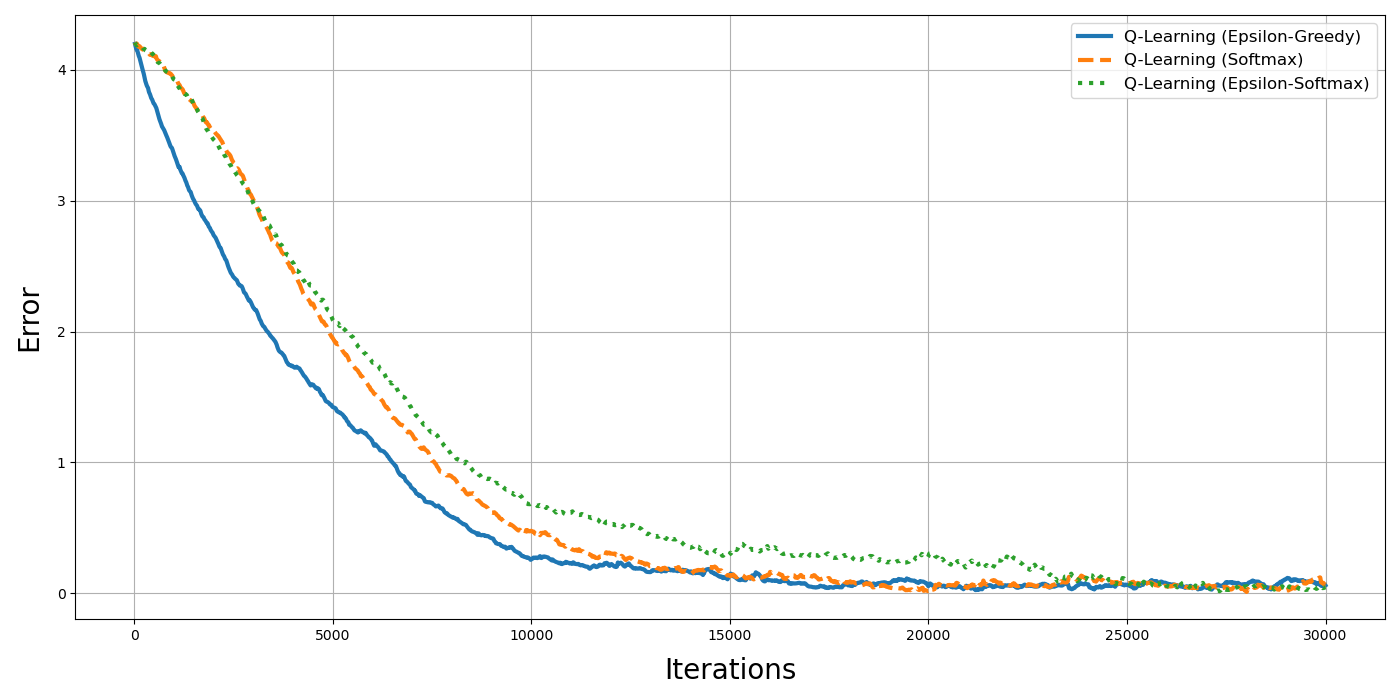}   \\
     \end{tabular}
  \end{center}
  \caption{Q-learning: Example with circular dynamic model }
  \label{fig:QLearning_01_0102}
\end{figure}

In this example we used the discount factor $\beta = 0.9$, step-size $\alpha = 0.01$, number of iterations $T_{\max}= 30000$ and exploration factor  $\epsilon = 0.3.$

From Fig.~\ref{fig:QLearning_01_0102a}, we can observe that the different modifications of Q-Learning such as: Q-Learning with $\epsilon$-greedy policy, Q-Learning with softmax policy, and Q-Learning with $\epsilon$-softmax policy are able to learn the true state values.

Fig.~\ref{fig:QLearning_01_0102} shows that the error in the  value function, i.e., $\Delta V_t$ falls steeply in the first $10000$ iterations for all algorithms. Additionally, we can observe that Q-Learning with $\epsilon$-greedy policy has better convergence rate than softmax and $\epsilon$-softmax.

\subsubsection{Example with no structure on transition model}
This example has five states and there is no structural assumption in the transition probability matrices. However, there is one for the reward matrix. For passive action ($a = 0$), the reward is linearly increasing along the states, and for active action ($a = 1$), the reward is linearly decreasing along the states.
\begin{eqnarray*}
  P_0 &=&  \begin{bmatrix}
    0.1502 & 0.0400 & 0.4156 & 0.0300 & 0.3642 \\
    0.4000 & 0.3500& 0.0800& 0.1200& 0.0500 \\
    0.5276 & 0.0400 &0.3991 & 0.0200 & 0.0133  \\
    0.0500 & 0.1000& 0.1500& 0.2000& 0.5000 \\
    0.0191 & 0.0100 & 0.0897 & 0.0300 & 0.8512 
\end{bmatrix}, \\
 P_1 &=& 
\begin{bmatrix}
    0.7196 & 0.0500 & 0.0903 & 0.0100 & 0.1301 \\
    0.5500 & 0.2000& 0.0500& 0.0800& 0.1200 \\ 
    0.1903 & 0.0100 & 0.1663 & 0.0100 & 0.6234 \\
    0.2000 & 0.0500 &0.1500 & 0.1000 & 0.5000 \\
    0.2501 & 0.0100& 0.3901 & 0.0300 & 0.3198 
\end{bmatrix}, \\
R &= &
\begin{bmatrix}
    0.4580 & 0.9631  \\
    0.5100 & 0.8100 \\
    0.6508 & 0.7963  \\
    0.6710 & 0.6061 \\
    0.6873 & 0.5057  
\end{bmatrix}
\end{eqnarray*}

In this example we used the discount factor $\beta = 0.9$, step-size $\alpha = 0.02$, number of iterations $T_{\max}= 30000$ and exploration factor  $\epsilon = 0.3.$

\begin{figure}
  \begin{center}
    \begin{tabular}{cc}
      \includegraphics[scale=0.23]{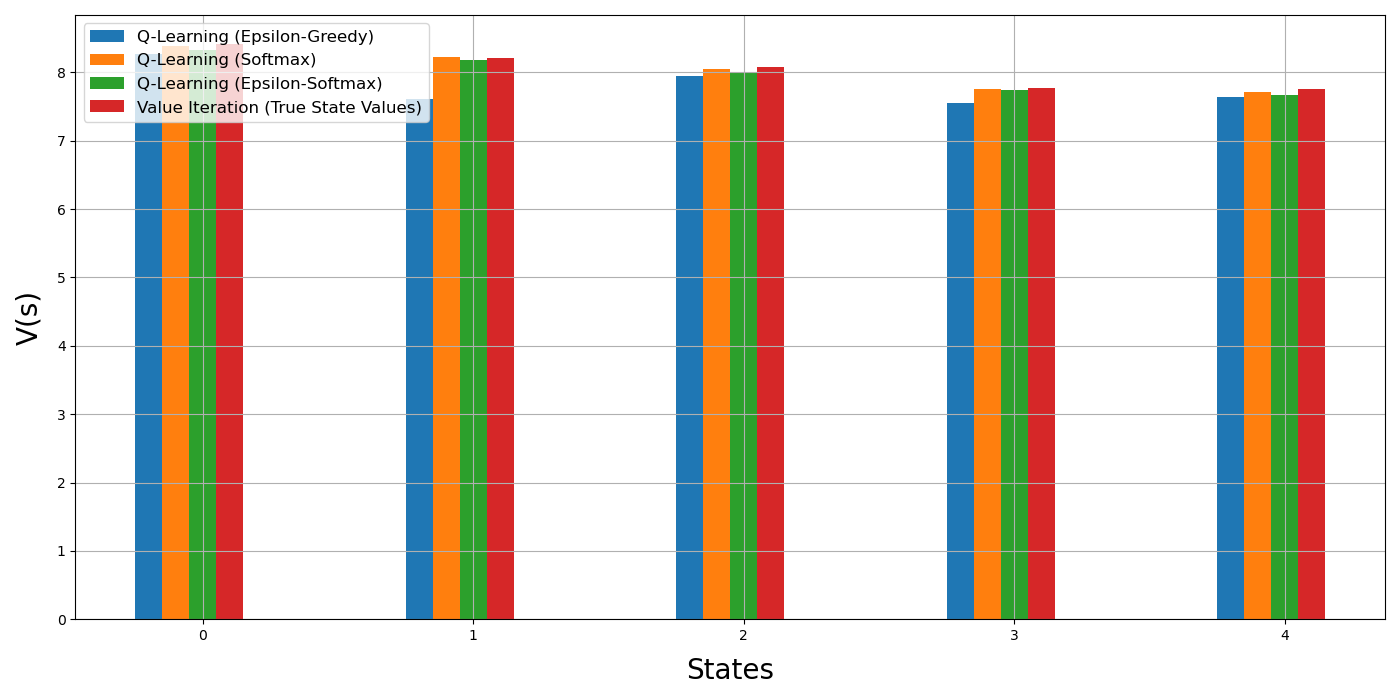}
    \end{tabular}
  \end{center}
  \caption{Q-learning: Example with no structure on transition model}
  \label{fig:QLearning_02_0102a}
\end{figure} 

\begin{figure}
  \begin{center}
    \begin{tabular}{cc}
      \includegraphics[scale=0.23]{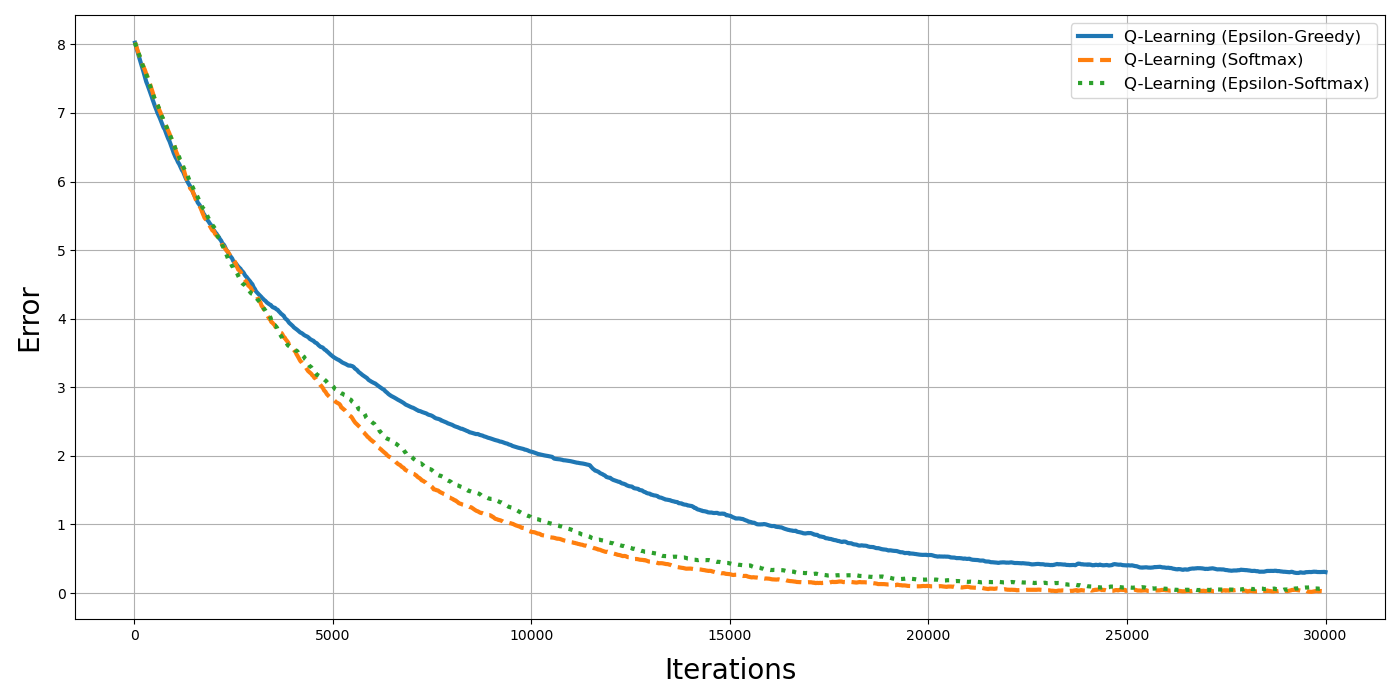}  \\
     \end{tabular}
  \end{center}
  \caption{Q-learning: Example with no structure on transition model}
  \label{fig:QLearning_02_0102}
\end{figure} 

From Fig.~\ref{fig:QLearning_02_0102a}, we can observe that the different modifications of Q-Learning such as: Q-Learning with $\epsilon$-greedy policy, Q-Learning with softmax policy, and Q-Learning with $\epsilon$-softmax policy are able to learn the true state values. Fig.~\ref{fig:QLearning_02_0102} shows that the error in the  values, i.e., $\Delta V_t$ falls steeply in the first $15000$ iterations for all  algorithms. Additionally, we can observe that Q-Learning with softmax and $\epsilon$-softmax policies are  better convergence rate than the $\epsilon$-greedy policy.

\subsubsection{Example with restart model}
 This example is taken from \cite{Avrachenkov2022}. It has five states and there are structural assumptions for both, transition probability and reward matrices. The transition probability matrices have the following structure: if action $1$ is taken, then $p^{1}_{s,1} = 1$ for all $s \in \mathcal{S}$.
\begin{eqnarray*}
  P_0 &=&  \begin{bmatrix}
    1/10 & 9/10 & 0 & 0 & 0 \\
    1/10 & 0 & 9/10 & 0 & 0 \\
    1/10 & 0 & 0 & 9/10 & 0 \\
    1/10 & 0 & 0 & 0 & 9/10 \\
    1/10 & 0 & 0 & 0 & 9/10
\end{bmatrix}, \\
 P_1 &=& 
\begin{bmatrix}
    1 & 0 & 0 & 0 & 0 \\
    1 & 0 & 0 & 0 & 0 \\
    1 & 0 & 0 & 0 & 0 \\
    1 & 0 & 0 & 0 & 0 \\
    1 & 0 & 0 & 0 & 0
\end{bmatrix}.
\end{eqnarray*}

The rewards for  passive action $(a = 0)$ with state $k,$ $r(k,0) = 0.9^k$ and the rewards for active action  $(a = 1)$, $r(k,1) =0$ for all $k \in \mathcal{S}$.

Here, we consider the discount factor $\beta = 0.9$, step-size $\alpha = 0.025$, number of iterations $T_{\max}= 30000$ and exploration factor  $\epsilon = 0.3.$

\begin{figure}
  \begin{center}
    \begin{tabular}{cc}
      \includegraphics[scale=0.23]{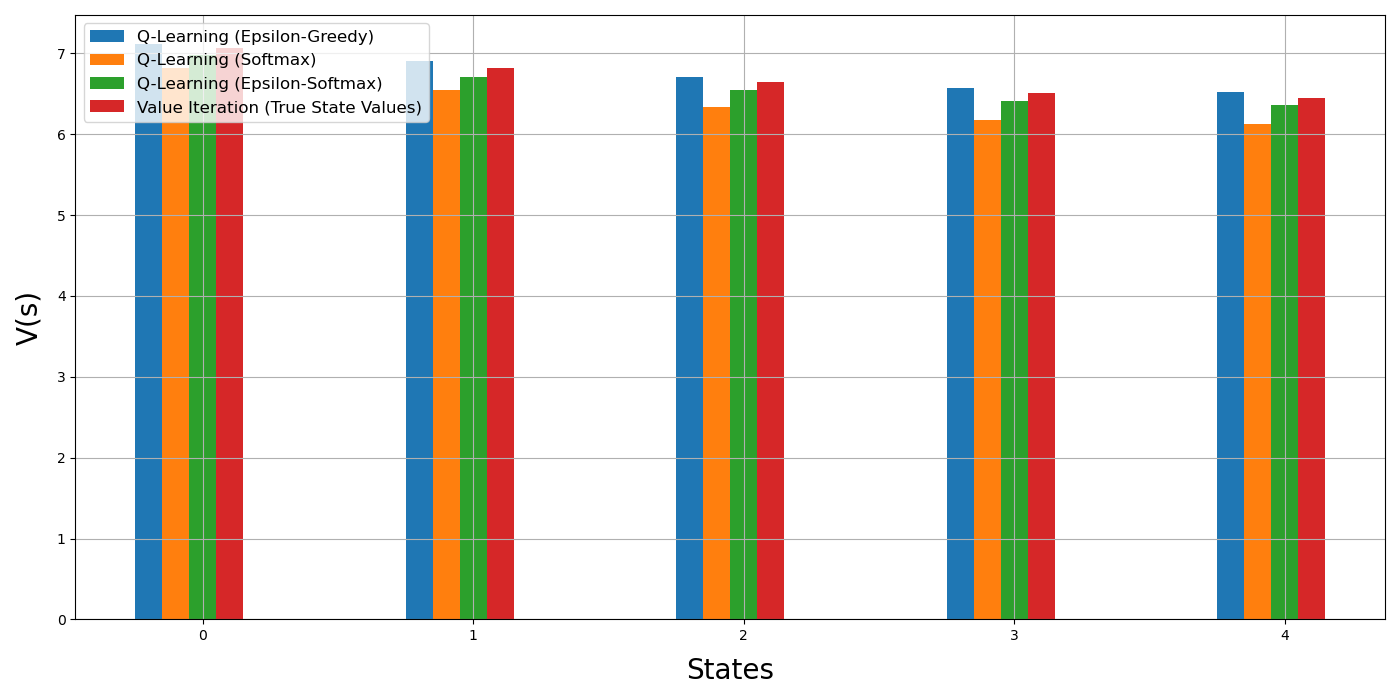}
     \end{tabular}
  \end{center}
  \caption{Q-learning: Example with restart model}
  \label{fig:QLearning_03_0102a}
\end{figure}

\begin{figure}
  \begin{center}
    \begin{tabular}{cc}
      \includegraphics[scale=0.23]{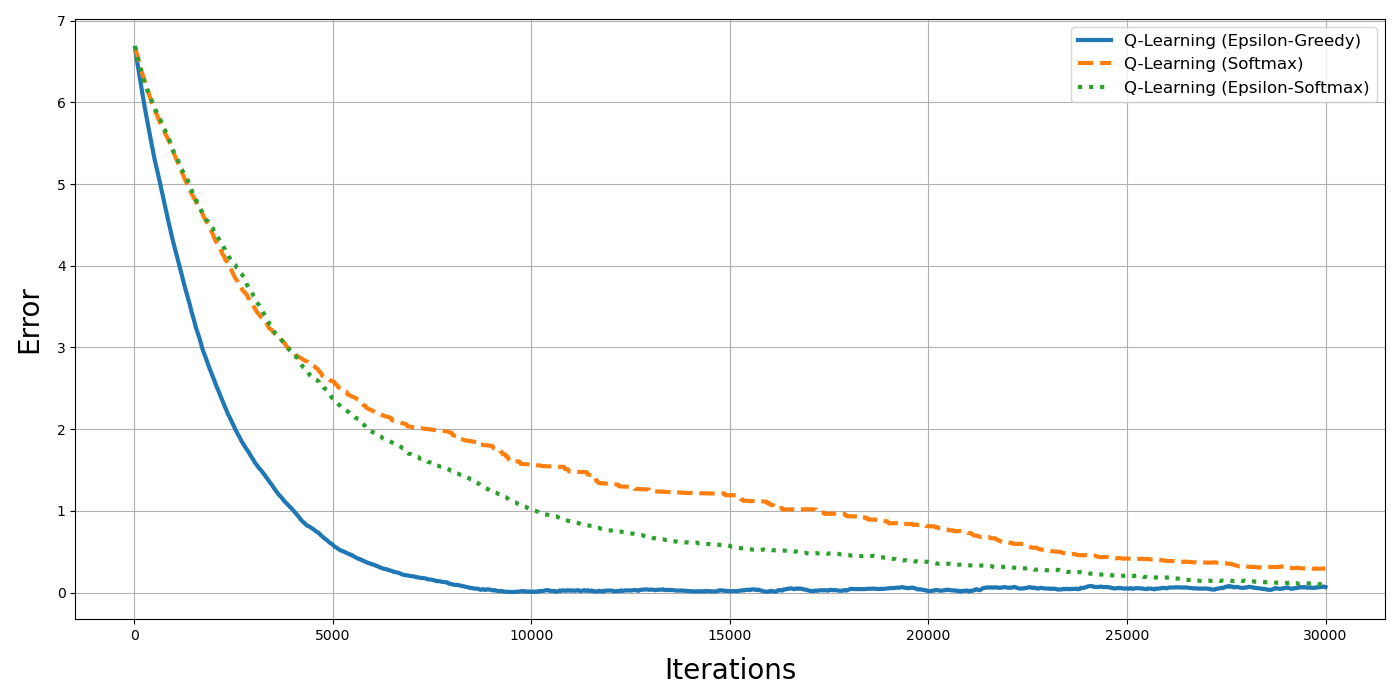}  \\
     \end{tabular}
  \end{center}
  \caption{Q-learning: Example with restart model}
  \label{fig:QLearning_03_0102}
\end{figure}

From Fig.~\ref{fig:QLearning_03_0102a}, we can observe that the different modifications of Q-Learning such as: Q-Learning with $\epsilon$-greedy policy, Q-Learning with softmax policy, and Q-Learning with $\epsilon$-softmax policy are able to learn the true state values. Fig.~\ref{fig:QLearning_03_0102} shows that the error in the state values, i.e., $\Delta V_t$ falls steeply in the first $15000$ iterations for all  algorithms. We notice that  $\epsilon$-greedy policy has higher convergence rate than softmax and $\epsilon$ softmax policy. 

\subsubsection{Example of one-step random walk with $K = 5$}
The probability matrix is same for both the actions $a = 1$ and $a = 0$. Rewards for passive action $r(k,0) = 0$ and active action $r(k,1) = 0.9^k$ for all $k \in \mathcal{S}$. This is also applicable in wireless communication systems. The probability matrices are given for $K = 5$ states.
\begin{eqnarray*}
  P_0 &=&  \begin{bmatrix}
    3/10 & 7/10 & 0 & 0 & 0 \\
    1/10 & 2/10 & 7/10 & 0 & 0 \\
    0 & 1/10 & 2/10 & 7/10 & 0 \\
    0 & 0 & 1/10 & 2/10 & 7/10 \\
    0 & 0 & 0 & 3/10 & 7/10
\end{bmatrix}, \\
 P_1 &=& P_0.
\end{eqnarray*}
%

We used   $\beta = 0.9$, $\alpha = 0.1$, $T_{\max}= 100000$ and exploration factor  $\epsilon = 0.3.$

\begin{figure}
  \begin{center}
    \begin{tabular}{cc}
      \includegraphics[scale=0.23]{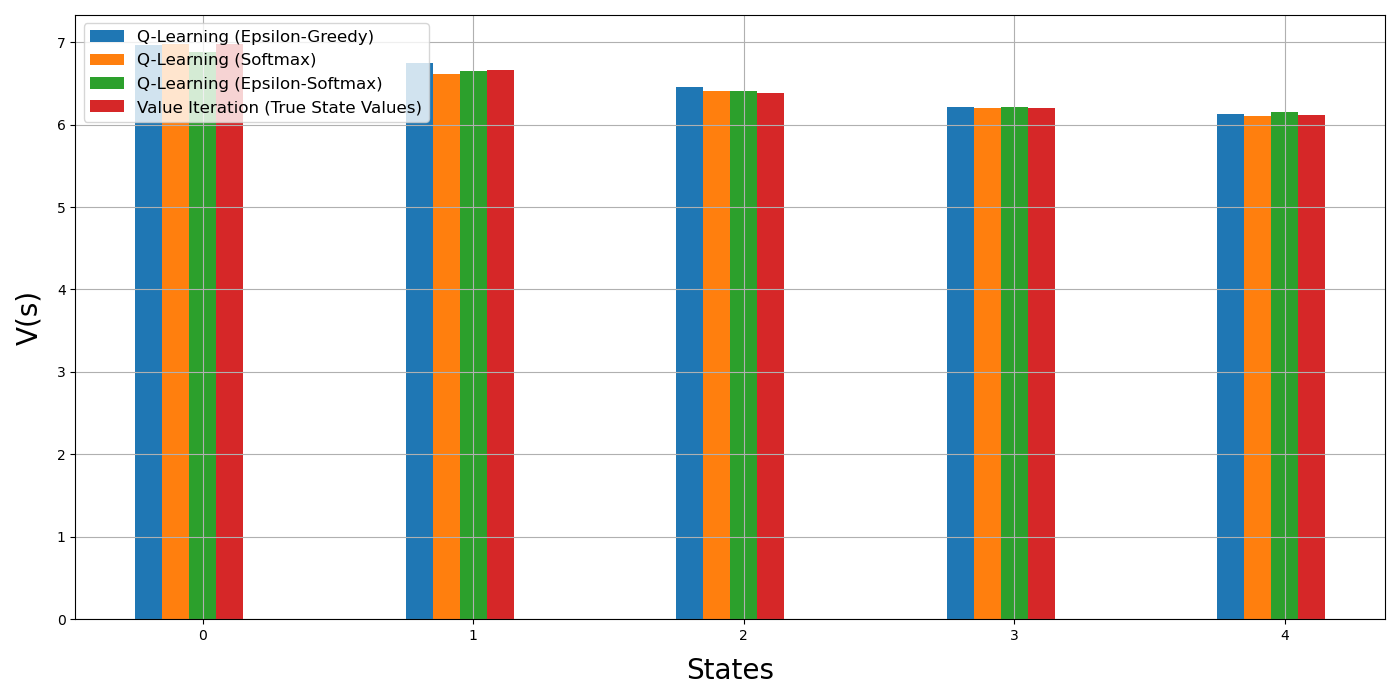}
     \end{tabular}
  \end{center}
  \caption{Q-learning: Example with one step random walk, number of states $K=5$}
  \label{fig:QLearning_04_0102a}
\end{figure}

\begin{figure}
  \begin{center}
    \begin{tabular}{cccc}
      \includegraphics[scale=0.23]{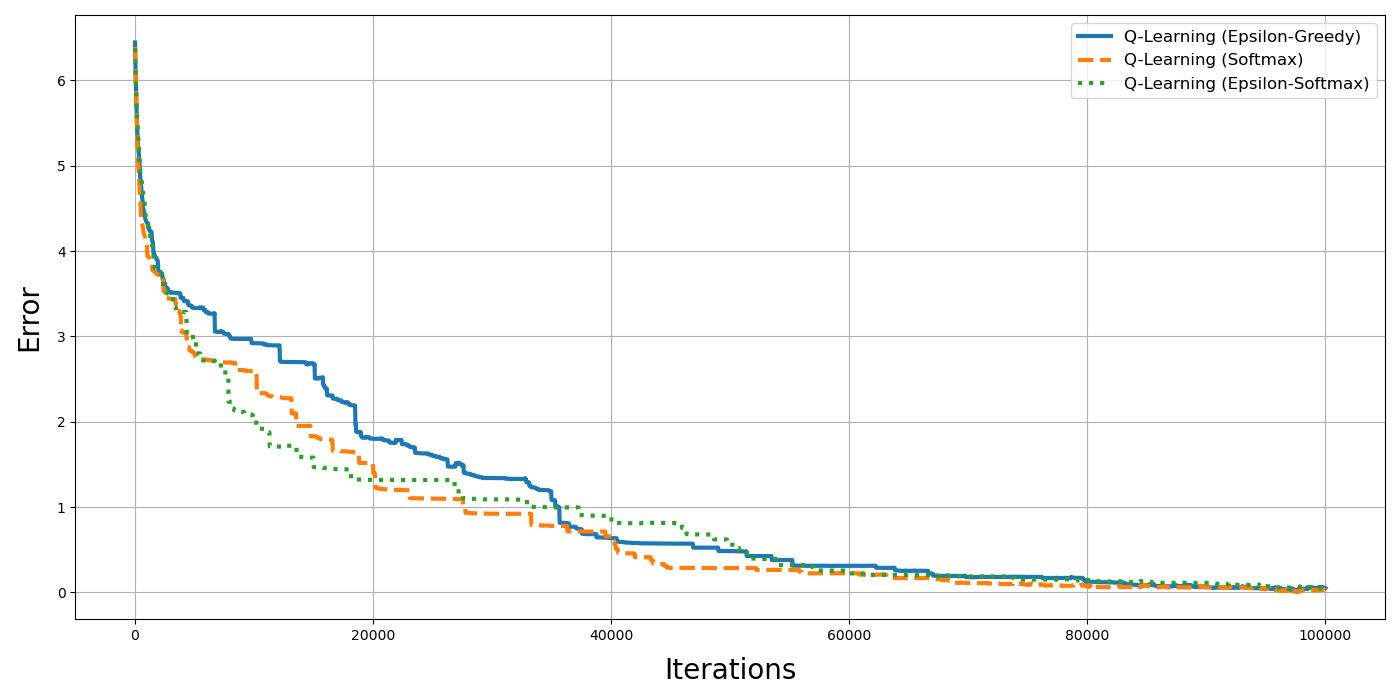}  \\
     \end{tabular}
  \end{center}
  \caption{Q-learning: Example with one step random walk, number of states $K=5$}
  \label{fig:QLearning_04_0102}
\end{figure}

From Fig.~\ref{fig:QLearning_04_0102a}, we can observe that the different modifications of Q-Learning such as: Q-Learning with $\epsilon$-greedy policy, Q-Learning with softmax policy, and Q-Learning with $\epsilon$-softmax policy are able to learn the true state values. Fig.~\ref{fig:QLearning_04_0102} shows that the error in the state values, i.e., $\Delta V_t$ falls steeply in the first $40000$ iterations for all  algorithms. Additionally, we can observe that Q-Learning with softmax and $\epsilon$-softmax policies are  better than the $\epsilon$-greedy policy in the initial iterations. 

\subsubsection{Example of one-step random walk with $K = 25$}

\begin{figure}
  \begin{center}
    \begin{tabular}{cccc}
     \includegraphics[scale=0.20]{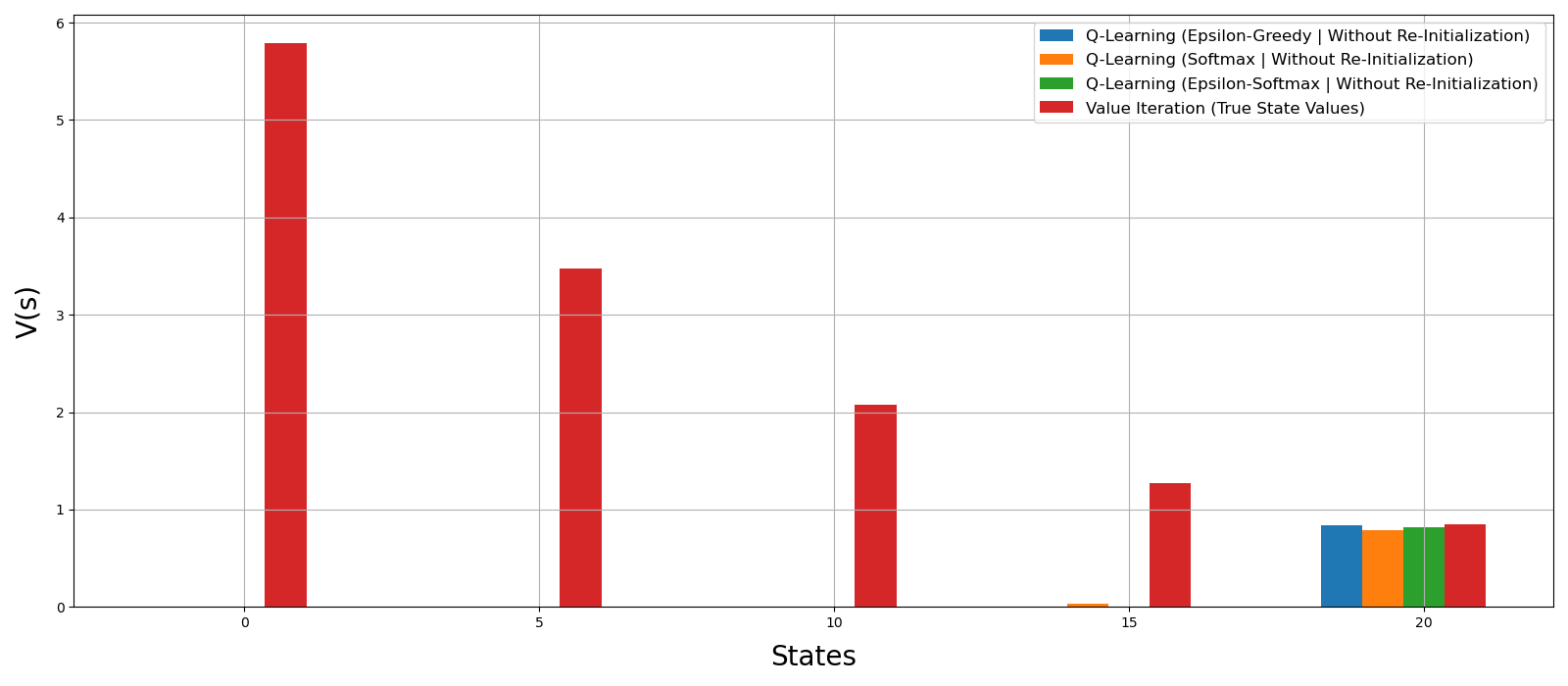}
     \end{tabular}
  \end{center}
  \caption{Q-learning: Example with one step random walk and number of states $K=25$}
  \label{fig:QLearning_05_01_0102a}
\end{figure}

\begin{figure}
  \begin{center}
    \begin{tabular}{cccc}
      \includegraphics[scale=0.20]{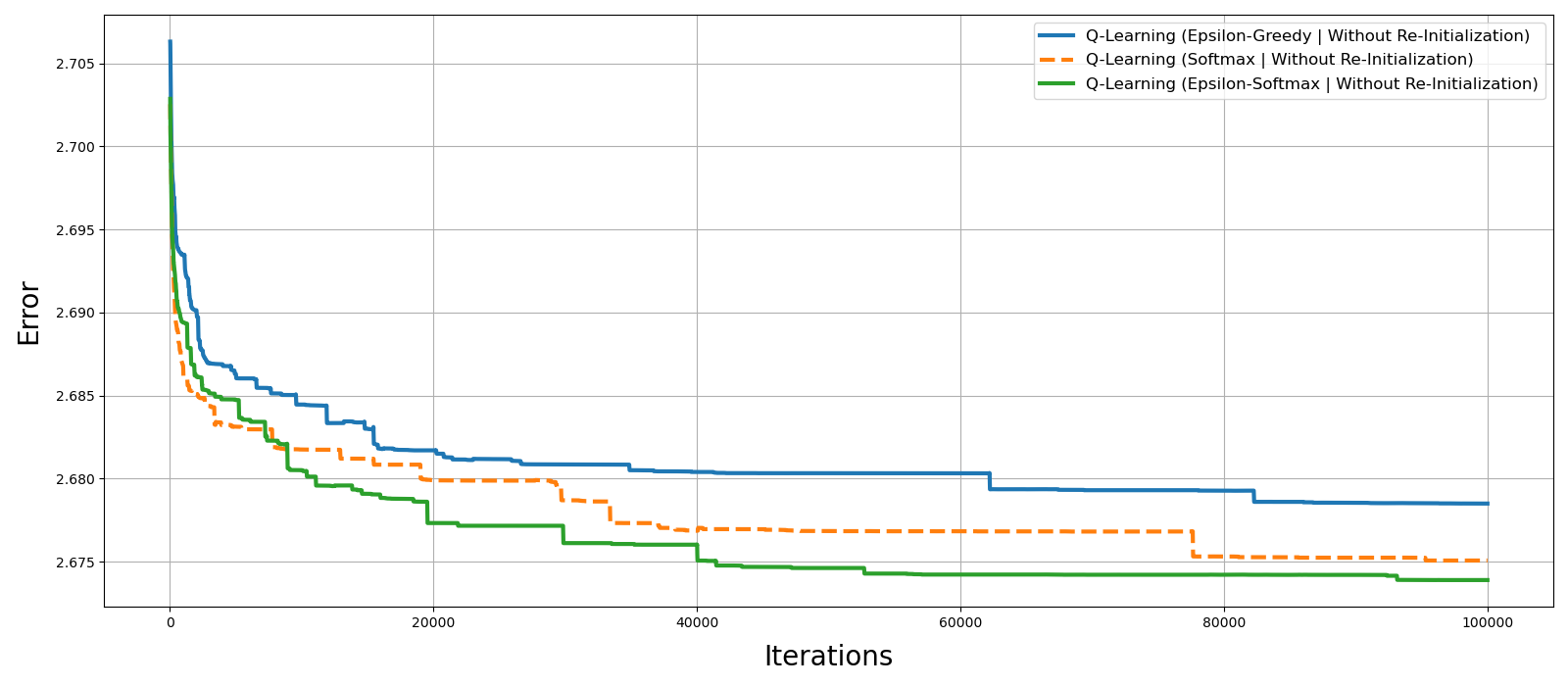}   \\
     \end{tabular}
  \end{center}
  \caption{Q-learning: Example with one step random walk and number of states $K=25$}
  \label{fig:QLearning_05_01_0102}
\end{figure}

Similar to the one-step random walk model with $K = 5$ states, we create probability matrices for $K = 25$ states. Since the states are larger in number compared to previous examples and transition model can affect the convergence of Q-learning with $\epsilon$-greedy, softmax and $\epsilon$-softmax. This may be due likelihood of transition into higher state, and less frequently into the lower state value. Those state-action value updates can be less frequent.  The convergence can be seen from Fig.~\ref{fig:QLearning_05_01_0102a}.

From Fig.~\ref{fig:QLearning_05_01_0102}, we can observe that none of the modifications of Q-Learning such as: Q-Learning with $\epsilon$-greedy policy, Q-Learning with softmax policy, and Q-Learning with $\epsilon$-softmax policy are able to learn the true state values. States which Q value function is not updated, they are having value $0$ and hence error is not converging to $0$ for all Q-learning algorithms.  Fig.~\ref{fig:QLearning_05_01_0102}-b shows that the error in the state values, i.e., $\Delta V_t$ only falls by a small level, and stabilizes afterwards.
We used parameter values $\beta = 0.9$, $\alpha = 0.2$, $T_{\max}= 100000$ and exploration factor  $\epsilon = 0.4.$ 

\subsubsection{Example of one-step random walk with $K = 25$ and re-initialization}


We use the same model as the previous one, the only difference being the re-initialization of the current state. The re-initialization of the current state is done in a random manner (reset the state  to random value of state after $50$ iterations to further allow exploration). This is a similar to episodic variant of Q learning. 

\begin{figure}
  \begin{center}
    \begin{tabular}{cccc}
     \includegraphics[scale=0.20]{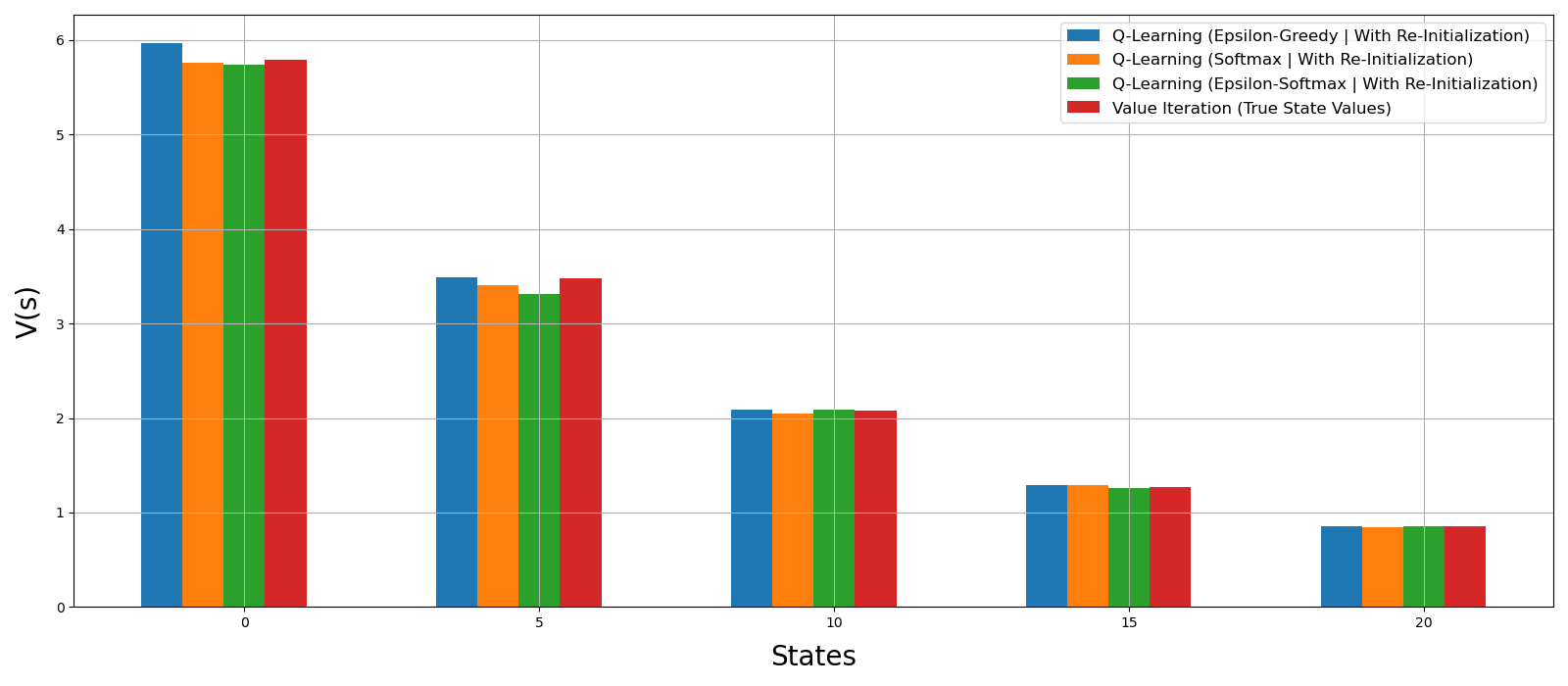}
     \end{tabular}
  \end{center}
  \caption{Q-learning: Example with one step random walk, number of state $K=25$ and with re-initialization}
  \label{fig:QLearning_05_02_0102a}
\end{figure}

\begin{figure}
  \begin{center}
    \begin{tabular}{cccc}
      \includegraphics[scale=0.20]{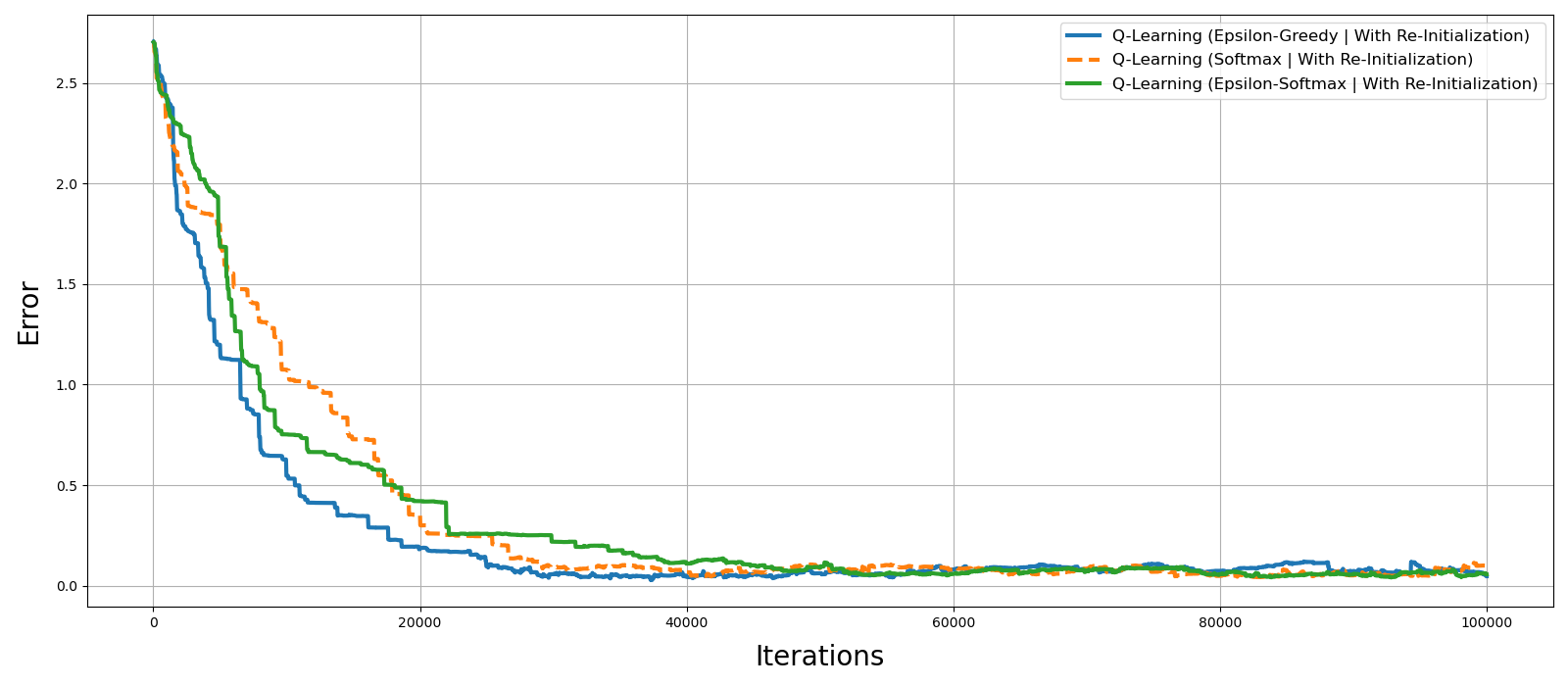}   \\
     \end{tabular}
  \end{center}
  \caption{Q-learning: Example with one step random walk, number of state $K=25$ and with re-initialization}
  \label{fig:QLearning_05_02_0102}
\end{figure}

From Fig.~\ref{fig:QLearning_05_02_0102a}, we can observe that the different modifications of Q-Learning such as: Q-Learning with $\epsilon$-greedy policy, Q-Learning with softmax policy, and Q-Learning with $\epsilon$-softmax policy are able to learn the true state values. Fig.~\ref{fig:QLearning_05_02_0102} shows that the error in the state values, i.e., $\Delta V_t$ falls steeply in the first $20000$ iterations. In these simulations, after every $50$ iterations, the current state is re-initialized. The rest of the hyper-parameter configuration is kept the same as before. 

\subsection{Numerical examples for index learning in SARB using Q-learning } 

We plot $\Delta \lambda_k$ as function of $k$ iteration. We study  examples as discussed in preceding section for index learning.

\begin{figure*}
  \begin{center}
    \begin{tabular}{cccc}
     \includegraphics[scale=0.23]{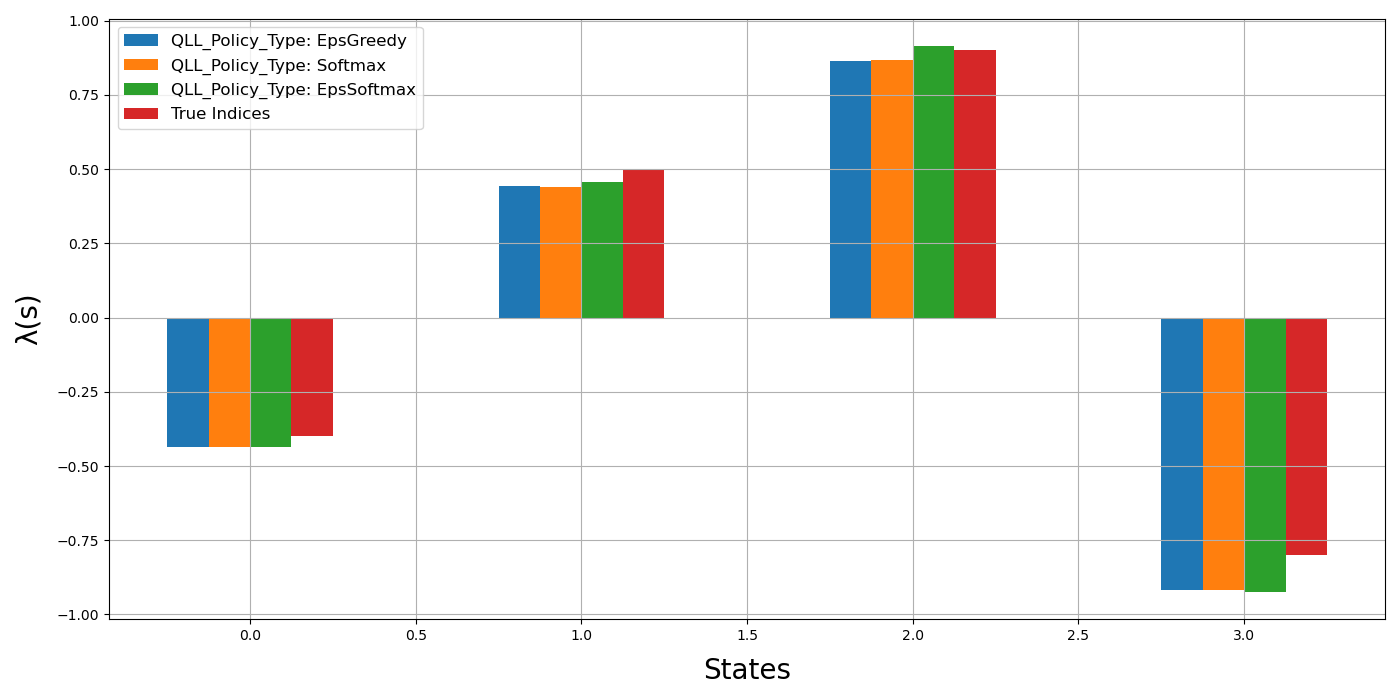}
      & 
      \includegraphics[scale=0.23]{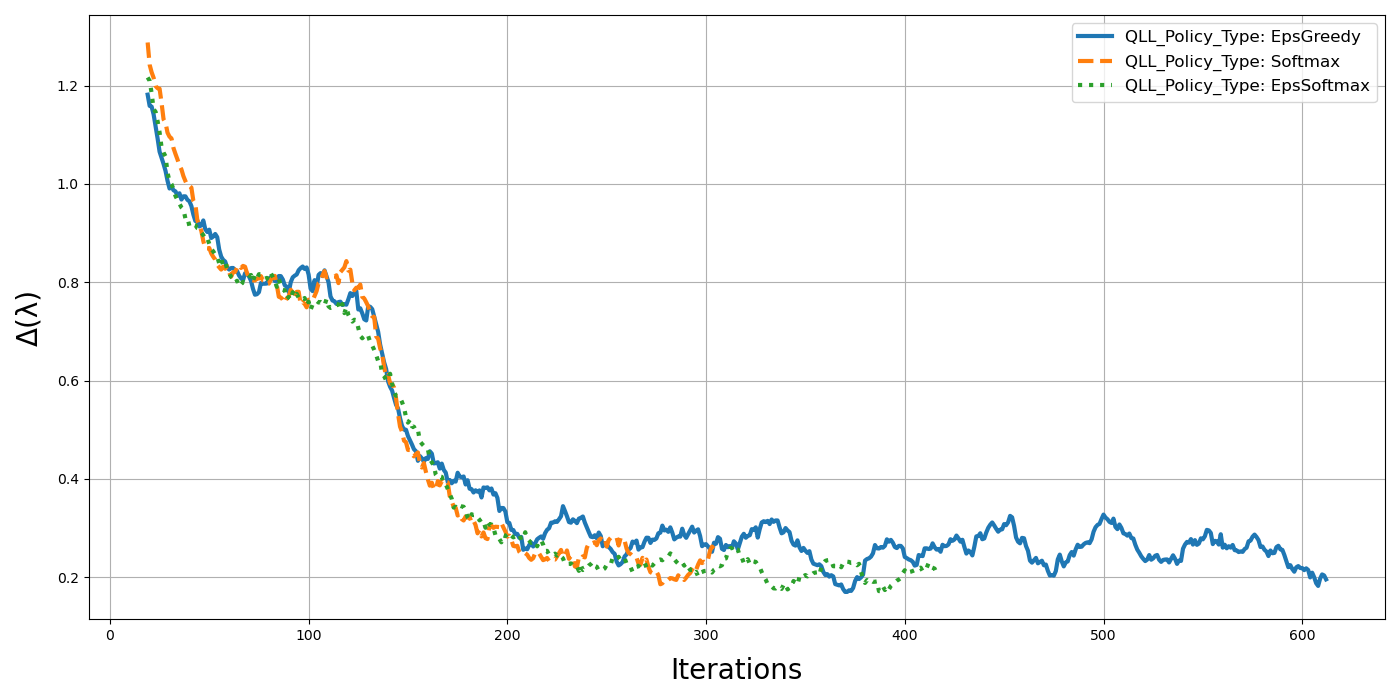}   \\
      a)  index values vs state & b) error vs iteration
     \end{tabular}
  \end{center}
  \caption{Index learning using Q learning: Example with circular dynamics model}
  \label{fig:QLLNR_01_0102}
\end{figure*}

\begin{figure*}
  \begin{center}
    \begin{tabular}{cccc}
     \includegraphics[scale=0.23]{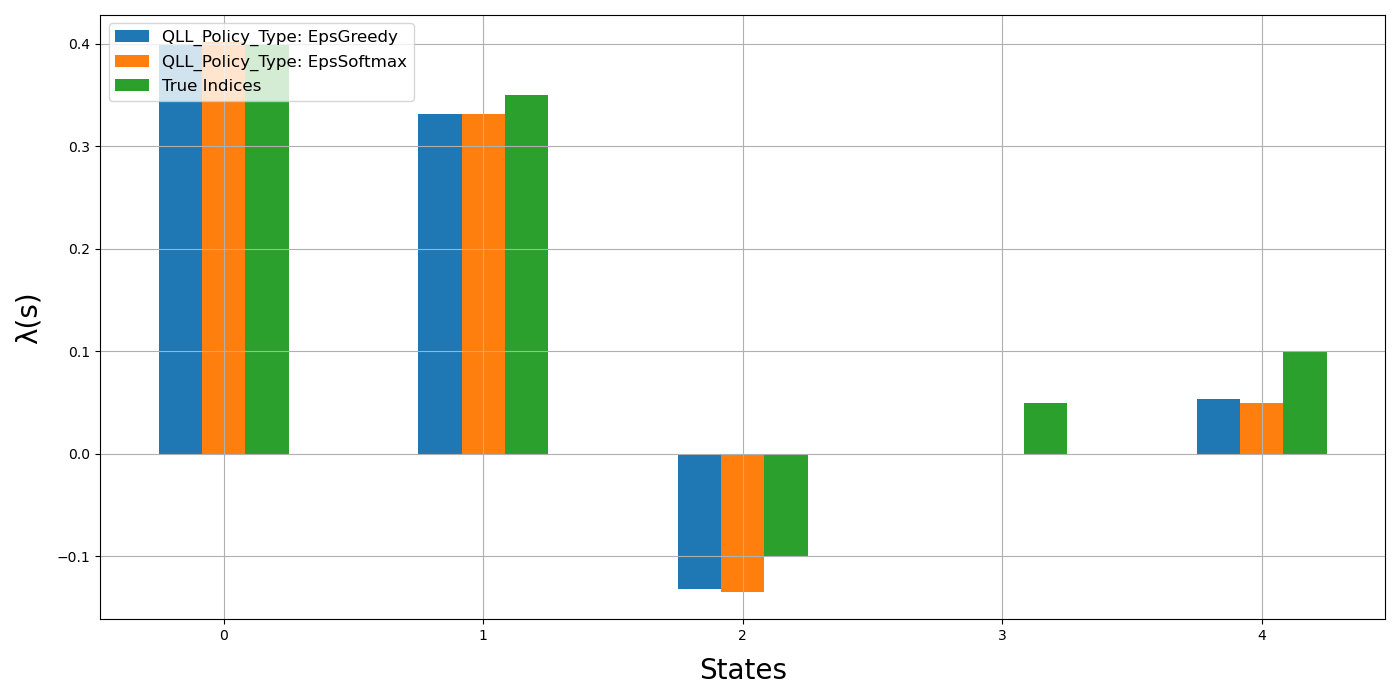}
      & 
      \includegraphics[scale=0.23]{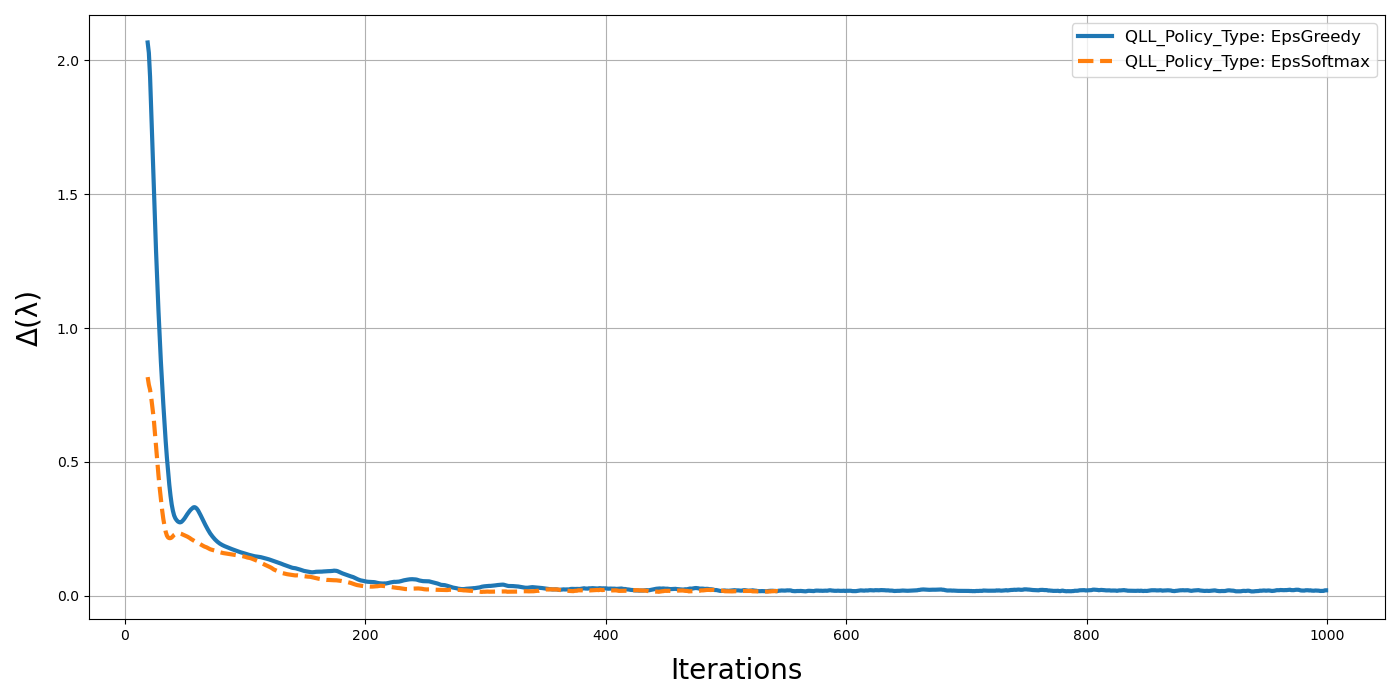}   \\
      a)  index values vs state & b) error vs iteration
     \end{tabular}
  \end{center}
  \caption{Index learning using Q learning: Example with   no structure on transition model}
  \label{fig:QLLNR_02_0102}
\end{figure*}

\begin{figure*}
  \begin{center}
    \begin{tabular}{cccc}
     \includegraphics[scale=0.23]{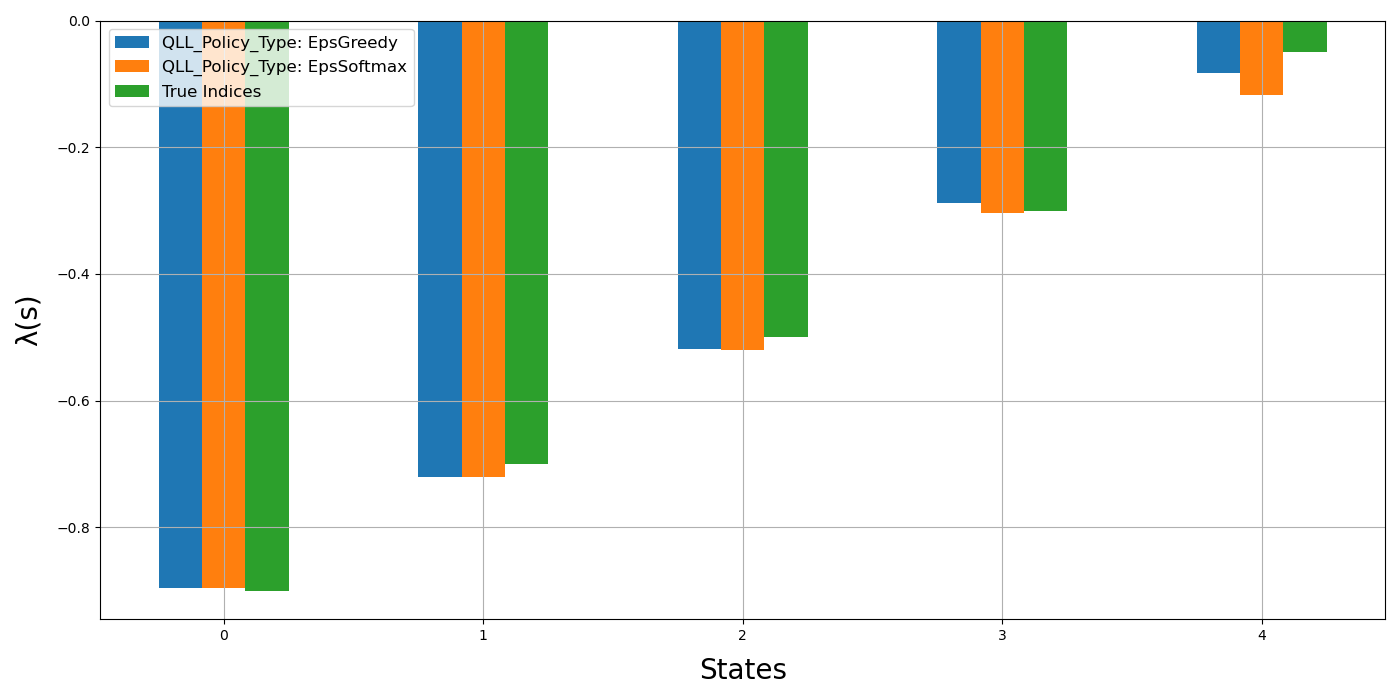}
      & 
      \includegraphics[scale=0.23]{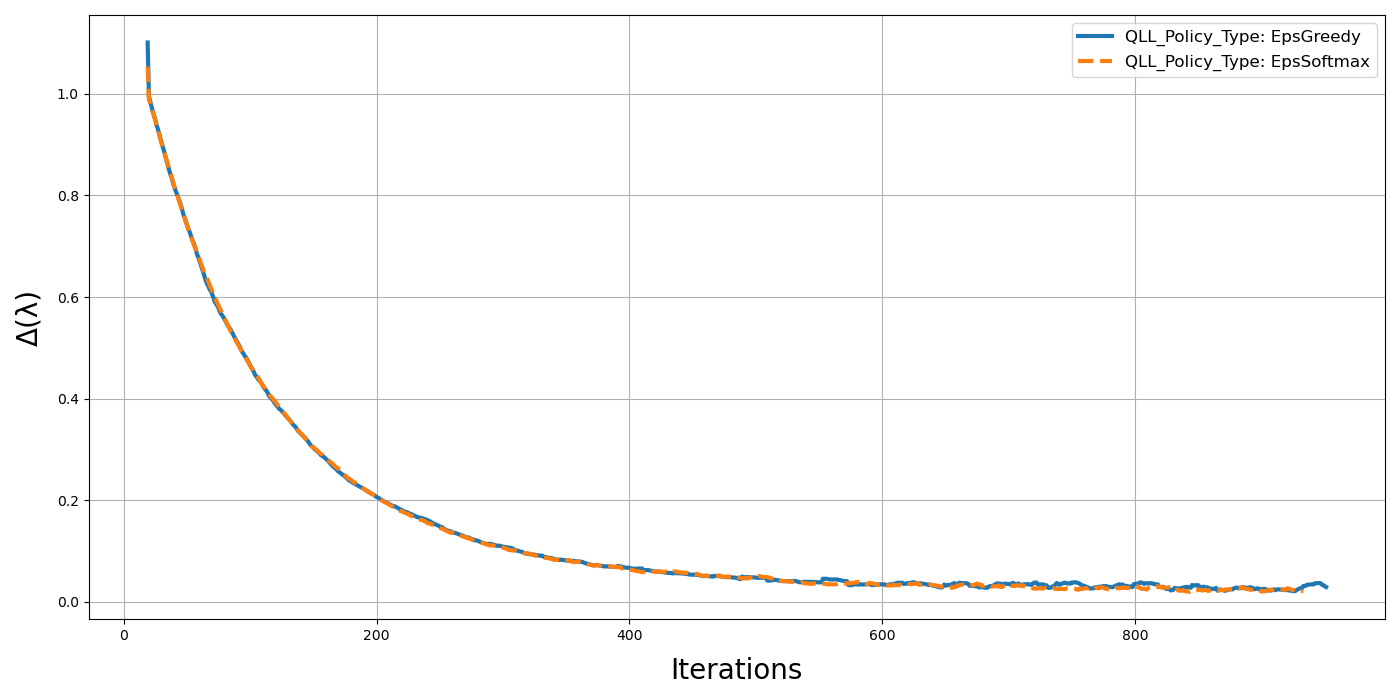}   \\
      a)  index values vs state & b) error vs iteration
     \end{tabular}
  \end{center}
  \caption{Index learning using Q learning: Example with  Restart Model }
  \label{fig:QLLNR_03_0102}
\end{figure*}

\begin{figure*}
  \begin{center}
    \begin{tabular}{cccc}
     \includegraphics[scale=0.23]{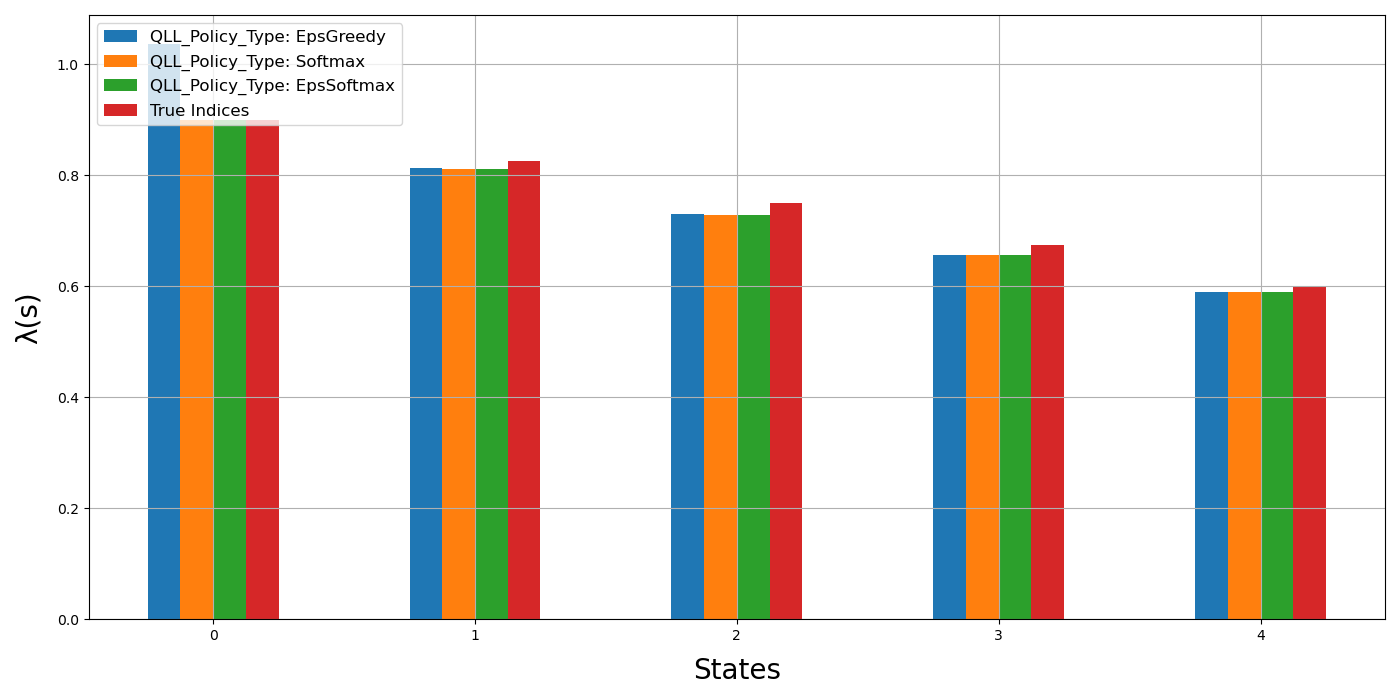}
      & 
      \includegraphics[scale=0.23]{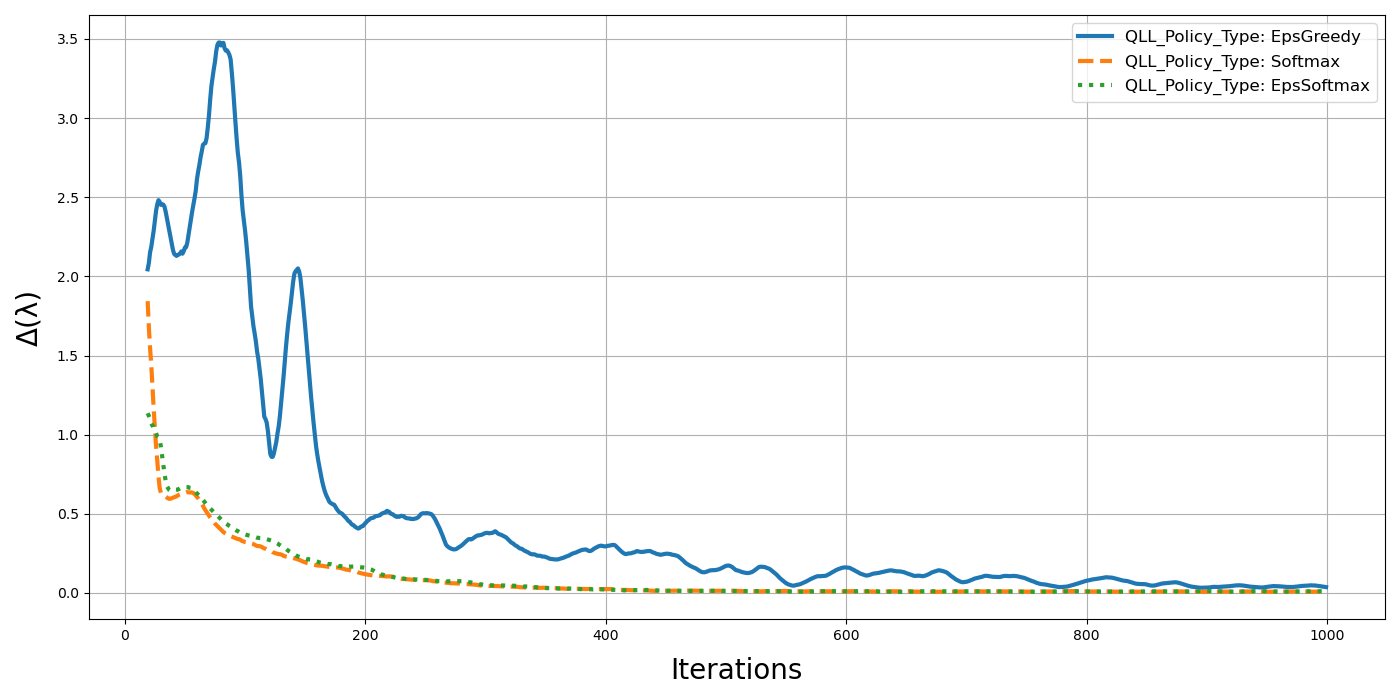}   \\
      a)  index values vs state & b) error vs iteration
     \end{tabular}
  \end{center}
  \caption{Index learning using Q learning: Example with  One step random walk with $K=5$  }
  \label{fig:QLLNR_04_0102}
\end{figure*} 

\subsubsection{Example with circular dynamics}
In this example, we use same parameter that discussed in preceding section example of circular dynamics.

Fig.~\ref{fig:QLLNR_01_0102}-b shows that the error in the indices, i.e., $\Delta \lambda_k$ falls steeply in the first 200 iterations for index learning using Q-learning with all the three policies: $\epsilon$-greedy, softmax, and $\epsilon$-softmax. Furthermore, from Fig.~\ref{fig:QLLNR_01_0102}, we can observe that index learning with all the three policies is able to learn the true indices. The  hyper-parameter configurations are as follows: $\beta = 0.9$, $\epsilon = 0.4$, $K_{\max} = 1000$, $T_{\max} = 5000$, $\alpha = 0.05$, $\gamma = 0.01$, $\delta = 0.05$.
\subsubsection{Example with no structure on transition model}

We consider same parameter as discussed in preceding section for this model. 

Fig.~\ref{fig:QLLNR_02_0102} shows that the error in the indices, i.e., $\Delta \lambda_k$ falls steeply in the first 200 iterations for index learning with the two policies: $\epsilon$-greedy and $\epsilon$-softmax. Furthermore, from Fig.~\ref{fig:QLLNR_02_0102}, we can observe that Q-$\lambda$ learning with the two policies is able to learn the true indices. Note that the softmax policy performs poorly for this model, hence, it has been excluded from the comparisons. The rest of the hyper-parameter configurations are as follows: $\beta = 0.9$, $\epsilon = 0.4$, $K_{\max} = 1000$, $T_{\max} = 5000$, $\alpha = 0.05$, $\alpha = 0.01$, $\delta = 0.005.$ 

%
\subsubsection{Example with restart model}
This section contains results for the example in B-3.

Fig.~\ref{fig:QLLNR_03_0102}-b shows that the error in the indices, i.e., $\Delta \lambda_k$ falls steeply in the first 400 iterations for Q-$\lambda$ learning with the two policies: $\epsilon$-greedy and $\epsilon$-softmax. Furthermore, from Fig.~\ref{fig:QLLNR_03_0102}, we can observe that Q-$\lambda$ learning with the two policies is able to learn the true indices. Note that the softmax policy performs poorly for this MDP, hence, it has been excluded from the comparisons. The rest of the hyper-parameter configurations are as follows: $\beta = 0.9$, $\epsilon = 0.4$, $K_{\max} = 1000$, $T_{\max} = 5000$, $\alpha = 0.05$, $\gamma = 0.01$, $\delta = 0.005.$
\subsubsection{Example of one-step random walk with $K = 5$}
This section contains results for the example in B-4.

Fig.~\ref{fig:QLLNR_04_0102} shows that the error in the indices, i.e., $\Delta \lambda_k$ falls steeply in the first 200 iterations for Q-$\lambda$ learning with all the three policies: $\epsilon$-greedy, softmax, and $\epsilon$-softmax. Furthermore, from Fig.~\ref{fig:QLLNR_04_0102}, we can observe that Q-$\lambda$ learning with all the three policies is able to learn the true indices. The rest of the hyper-parameter configurations are as follows: $\beta = 0.9$, $\epsilon = 0.4$, $K_{\max} = 1000$, $T_{\max} = 5000$, $\alpha = 0.05$, $\gamma = 0.01$, $\delta = 0.001.$
%

\bibliographystyle{IEEEbib}
\bibliography{its-references}

\begin{thebibliography}{10}

\bibitem{Borkar17b}
V.~S. Borkar, G.~S. Kasbekar, S.~Pattathil, and P.~Y. Shetty,
\newblock ``Opportunistic scheduling as restless bandits,''
\newblock {\em IEEE Transactions on Control of Network Systems}, vol. 5, Dec. 2018.

\bibitem{Xiong2023}
G.~Xiong, S.~Wang, J.~Li, and R.~Singh,
\newblock ``Whittle index based {Q}-learning for wireless edge caching with linear function approximation,''
\newblock {\em Arxiv}, pp. 1--16, Feb. 2023.

\bibitem{Verloop16}
I.~M. Verloop,
\newblock ``Asymptotically optimal priority policies for indexable and nonindexable restless bandits,''
\newblock {\em The Annals of Applied Probability}, vol. 26, no. 4, pp. 1947--1995, 2016.

\bibitem{Glazebrook05}
K.~D. Glazebrook, H.~M. Mitchell, and P.~S. Ansell,
\newblock ``Index policies for the maintenance of a collection of machines by a set of repairmen,''
\newblock {\em European Jorunal of Operation Research}, vol. 165, pp. 267--284, Aug. 2005.

\bibitem{Glazebrook2006}
D.~Ruiz-Hernandez K.~D.~Glazebrook and C.~Kirkbride,
\newblock ``Some indexable families of restless bandit problems,''
\newblock {\em Advances in Applied Probability}, vol. 38, no. 3, pp. 643--672, Sept. 2006.

\bibitem{Killian2021}
J.~A. Killian, A.~Biswas, S.~Shah, and M.~Tambe,
\newblock ``Q-learning {L}agrange policies for multi-action restless bandits,''
\newblock in {\em KDD-21}, 2021, vol.~27, pp. 871--881.

\bibitem{Mate21}
A.~Mate, A.~Perrault, and M.~Tambe,
\newblock ``Risk-aware interventions in public health: Planning with restless multi-armed bandits.,''
\newblock in {\em Proceedings of AAMAS}, 2021, pp. 880--888.

\bibitem{Meshram15}
R.~Meshram, D.~Manjunath, and A.~Gopalan,
\newblock ``A restless bandit with no observable states for recommendation systems and communication link scheduling,''
\newblock in {\em Proceedings of {CDC}}, Dec 2015, pp. 7820--7825.

\bibitem{Meshram2016}
R.~Meshram, A.~Gopalan, and D.~Manjunath,
\newblock ``Optimal recommendation to users that react: Online learning for a class of pomdps,''
\newblock in {\em Conference on Decision and Control}, 2016, vol.~55, pp. 1--6.

\bibitem{Meshram17}
R.~Meshram, A.~Gopalan, and D.~Manjunath,
\newblock ``A hidden {M}arkov restless multi-armed bandit model for playout recommendation systems,''
\newblock {\em COMSNETS, Lecture Notes in Computer Science, Springer}, pp. 335--362, 2017.

\bibitem{Ny2008}
Jerome~Le Ny, Munther Dahleh, and Eric Feron,
\newblock ``Multi-uav dynamic routing with partial observations using restless bandit allocation indices,''
\newblock in {\em ACC}, 2008, pp. 1--6.

\bibitem{Whittle88}
P.~Whittle,
\newblock ``Restless bandits: {A}ctivity allocation in a changing world,''
\newblock {\em Journal of Applied Probability}, vol. 25, no. A, pp. 287--298, 1988.

\bibitem{Gittins11}
J.~Gittins, K.~Glazebrook, and R.~Weber,
\newblock {\em Multi-armed bandit allocation indices},
\newblock Wiley, 2011.

\bibitem{NinoMora23}
J.~Nino Mora,
\newblock ``Markovian restless bandits and index policies: A review,''
\newblock {\em Mathematics}, 2023.

\bibitem{Weber90}
R.~R. Weber and G.~Weiss,
\newblock ``On an index policy for restless bandits,''
\newblock {\em Journal of Applied Probability}, vol. 27, no. 3, pp. 637--648, Sept. 1990.

\bibitem{Tekin2012}
C.~Tekin and M.~Liu,
\newblock ``Online learning of rested and restless bandits,''
\newblock {\em IEEE Transactions on Information Theory}, vol. 58, no. 8, pp. 5588--5611, May 2012.

\bibitem{Fu2019}
J.~Fu, Y.~Nazarthy, S.~Moka, and P.~G. Taylor,
\newblock ``Towards {Q}-learning the {W}hittle index for restless bandits,''
\newblock in {\em Proceedings of Australian and New Zealand Control Conference (ANZCC)}, Nov. 2019, pp. 249--254.

\bibitem{Jung2019}
Y.~H. Jung and A.~Tiwari,
\newblock ``Regret bounds for thomspon sampling in episodic restless bandit problems,''
\newblock in {\em NeurIPS}, 2019, vol.~33, pp. 1--10.

\bibitem{Wang2020}
S.~Wang, L.~Huang, and J.~C.~S. Liu,
\newblock ``Restless-{UCB} an efficient and low-complexity algorithm for online restless bandits,''
\newblock in {\em NeurIPS}, 2020, vol.~34, pp. 1--12.

\bibitem{Nakhleh2021}
K.~Nakhleh, S.~Ganji, P.~C. Hsieh, I-H. Hou, and S.~Shakkottai,
\newblock ``{NeurWIN}: Neural {W}hittle index network for restless bandits via deep {RL},''
\newblock in {\em NeurIPS}, 2021, vol.~35, pp. 1--12.

\bibitem{Akbarzadeh2022}
N.~Akbarzadeh and A.~Mahajan,
\newblock ``Conditions for indexability of restless bandits and $o(k^3)$ algorithm to compute {W}hittle index,''
\newblock {\em Advances in Applied Probability}, vol. 54, pp. 1164--1192, 2022.

\bibitem{Avrachenkov2022}
K.~E. Avrachenkov and V.~S. Borkar,
\newblock ``Whittle index based {Q}-learning for restless bandits with average reward,''
\newblock {\em Automatica}, vol. 139, pp. 1--10, May 2022.

\bibitem{Watkins1989}
C.~J. Watkins,
\newblock {\em Learning from delayed rewards},
\newblock Phd Thesis, 1989.

\bibitem{Borkar2000}
V.~S. Borkar and S.~P. Meyn,
\newblock ``The {O.D.E.} method for convergence of stochastic approximation and reinforcement learning,''
\newblock {\em SIAM J. Control Optimization}, vol. 38, no. 2, pp. 447--469, 2000.

\bibitem{Borkar08}
V.~S. Borkar,
\newblock {\em Stochastic approximation: a dynamical systems viewpoint},
\newblock Cambridge University Press, 2008.

\bibitem{Sutton2018}
R.~S. Sutton and A.~G. Barto,
\newblock {\em Reinforcement learning: An Introduction},
\newblock MIT Press, 2018.

\bibitem{Satinder2000}
S.~Singh, T.~J. Akkola, M.~T. Littman, and C.~Szepesvari,
\newblock ``Convergence results for single-step on-policy reinforcement learning algorithms,''
\newblock {\em Machine Learning}, , no. 39, pp. 287--308, 2000.

\bibitem{Jaakkola1993}
T.~Jaakkola, M.~I. Jordon, and S.~P. Singh,
\newblock ``Convergence of stochastic iterative dynamic programming algorithms,''
\newblock in {\em Proceedings of NIPs}, 1993, pp. 703--710.

\bibitem{Mnih2015}
V.~Mnih1, K.~Kavukcuoglu1, A.~A.~Rusu1 D.~Silver, and J.~Veness et. al.,
\newblock ``Human-level control through deep reinforcement learning,''
\newblock {\em Nature}, vol. 518, pp. 529--533, 2015.

\bibitem{Robledo2021}
F.~Robledo, V.~S. Borkar, U.~Ayesta, and K.~E. Avrachenkov,
\newblock ``{QWI}:{Q}-learning with {W}hittle index,''
\newblock {\em Performance Evaluation Review}, vol. 49, no. 2, pp. 47--49, Sept. 2021.

\bibitem{Robledo2022}
F.~Robledo, V.~S. Borkar, U.~Ayesta, and K.~Avrachenkov,
\newblock ``Tabular and deep learning of whittle index,''
\newblock {\em European Workshop of Reinforcement Learning}, pp. 1--15, Sept 2022.

\bibitem{Laxminarayanan17}
C.~Laxminarayanan and S.~Bhatnagar,
\newblock ``A staiblity criteria for two timescale stochastic approximation,''
\newblock {\em Automatica}, , no. 79, pp. 108--114, 2017.

\end{thebibliography}

\end{document}